\documentclass[11pt]{article}

\usepackage[preprint]{acl}

\usepackage{times}
\usepackage{latexsym}
\usepackage{booktabs}
\usepackage{makecell}
\usepackage{tikz}
\usetikzlibrary{positioning,calc}

\usepackage[T1]{fontenc}

\usepackage[utf8]{inputenc}
\usepackage{xurl}
\usepackage{hyperref}

\usepackage{microtype}

\usepackage{inconsolata}

\usepackage{graphicx}

\author{
 \textbf{Ted Underwood\textsuperscript{1}},
\textbf{Ziliang Qiu\textsuperscript{1}},
\textbf{Sarah Griebel\textsuperscript{1}},
\textbf{Laura K. Nelson\textsuperscript{2}},
\\
\textbf{Edwin Roland\textsuperscript{1}},
\textbf{Wenyi Shang\textsuperscript{3}},
\textbf{Matthew Wilkens\textsuperscript{4}}
\\
\small
 \textsuperscript{1}University of Illinois Urbana-Champaign,
 \textsuperscript{2}University of British Columbia,
 \\
 \small
  \textsuperscript{3}University of Missouri,
 \textsuperscript{4}Cornell University
\\
\small{
\textbf{Correspondence:} \href{mailto:tunder@illinois.edu}{tunder@illinois.edu}
 }
}

\title{Chronologic: Measuring Language Models' Ability to Represent the Past}

\begin{document}
\maketitle
\begin{abstract}
Language models are appealing tools for research on the past. But to trust the evidence a model provides, researchers need to know whether its responses fit the period represented. Validation is challenging, because this is not a task living people ordinarily perform, and because many questions have multiple correct answers. We use historical texts to develop a benchmark for a model's representation of English-language contexts 1831-1930, relying on pairwise comparisons to multiple ground truths and strong distractors to score the hardest questions in an appropriately graduated way. We find that generative tasks are harder than discriminative ones; in fact, reasoning models can typically discern the weakness of their own generated answers. While models pretrained exclusively on historical text lead the pack when evaluated by answer likelihood, they cannot compete with commercial models in free generation. None of the models we tested represent historical contexts in a fully persuasive way yet, but progress toward that goal is evident.
\end{abstract}

\section{Introduction}

Language models' dependence on written training data creates remarkable opportunities for social scientists and historians, who can now transform collections of text into interactive systems. Some behavioral scientists hope language models will reproduce the ``collective mentality of the historical people whose writings were used to build them'' \citep{varnum2024historical_llms}. Others hope to produce ``capable research assistants'' whose knowledge aligns with the period studied \citep{goettlich_et_al_2025_history_llms_blog}. 

The success of this project is far from assured. Training data is scarce, and the goal of historical modeling must be framed carefully: it is not clear that modeling texts will produce a model of the minds that wrote them \citep{poibeau2026historical}. Moreover, the written record captures only a biased sample of past behavior, and compressing documents into a model may further homogenize differences between them \citep{bisbee2023synthetic, boelaert2024machine}. Models trained on the past might still struggle to speak from any \emph{specific} vantage point.

We have designed a benchmark in part to address these open questions about models' utility in research. But Chronologic can also assess the risk that casual users will be misled if they ask a model ``what Victorians thought about'' a social problem. Existing benchmarks catch models that make errors in math or logic. But when students ask a model to speak for the past, there is now no easy way to check whether its output is accurate or caricatured.

The benchmark we offer here (Chronologic-EN-1.0) evaluates models both as research tools and as informal interlocutors. The conversational scene we envision is that a contemporary user has asked the model to respond to a question like a writer in a specified historical context---not a named individual, but ``an American fashion writer in the 1870s,'' or ``an Indian historian in the 1920s.'' Can a model produce an answer that is accurate, that fits the specified social perspective, and that matches period style? Since many valid answers may be possible, evaluation is challenging.  

\section{Related work}

The well-known Western biases of commercial language models have produced a lively debate about pluralistic alternatives. Researchers have produced databases that aim to measure cultural diversity 
 \citep{rao2024normad, shi2024culturebank} and even full-blown ontologies for the representation of culture \citep{zhang_et_al_2025_culturescope}. Historical models are allied to this initiative: one could say that they expand cultural pluralism in the direction of the past.
 
 But what is ``culture''? When used by sociologists, the word can cover a process that subsumes all learned practices---even, say, base-10 arithmetic \citep{foster2018culture_computation}. In discussing the biases of language models, by contrast, ``culture'' becomes narrower. When \citet{adilazuarda2024culture} reviewed more than 90 cultural databases and benchmarks, they found a heavy emphasis on ``emotions and values, food and drink, kinship terms, social etiquette.'' In short, ``culture'' is often seen from the outside, and equated with the differentia between groups. 
 
\citet{zhou2025culture} have argued that this understanding of culture produces ``indexical'' knowledge about group stereotypes, without teaching models how to behave like a member of the group or understand cultural practices from within. We find this critique so persuasive that we have avoided separating cultural difference from other aspects of language here. Instead, we approach Chronologic as a capability benchmark, which measures familiar targets like inference and language generation---as those tasks were understood in English-speaking contexts between 1831 and 1930. Math may not loom large in catalogs of cultural difference, for instance, but it too changes: people who reckoned in shillings had to use mixed-radix arithmetic.

While benchmarks should ultimately cover many languages and periods, we believe an attempt to cover all of human history at once would gravely underestimate the difficulty of the task. Different contexts will require different sorts of domain expertise. We offer Chronologic-EN, not as a full solution for historical representation, but as a template that illustrates the shape of the problem.

There are practical reasons to begin with the century before 1930. The archive of texts available for language-model training peaks in this period, because publication practices and IP laws intersect to make texts before 1931 readily available, especially in English  \citep{cargnelutti2025institutionalbooks10242b}. As a result, many recently-released historical models were trained on English text up to some point in the first third of the twentieth century \citep{levine2026talkie, luo2026pretraininghistorical}. A benchmark comparing them to frontier models will be immediately useful.

\section{Data collection}

One challenge in designing a benchmark for the past is that living humans cannot reliably express themselves in the language of a vanished era. Ground truth answers have to come from historical artifacts. Our team includes scholars who have studied the nineteenth and early twentieth centuries; their domain expertise was valuable in selecting texts and framing questions. But the answers to questions---with rare exceptions---are drawn from period texts themselves. 

Those sources included diaries, letters, newspapers, textbooks, trial records, fiction, and poetry, as well as nonfiction books about a wide range of subjects. Our selection process began with random selection from Institutional Books 1.0, followed by rejection of authors who are well-known and frequently quoted on the internet. We then supplemented our sample with a deliberate search for underrepresented perspectives, including books by women, the autobiography of a formerly enslaved man, and English-language books published in Bombay, Dublin, Madras, Peking, and Tokyo \citep{idi_institutional_books_1_2024}. The resulting collection of 187 texts is not a probability sample, but maximizes diversity of perspective.

The risk of using public ground truth, of course, is that---although we try to avoid well-known authors and works---commercial models may have seen some of the texts we use. To assess this risk, we measured memorization using a version of the \citet{chang2023speak} method. We selected 16 books from our corpus that contained proper nouns either fictional, or obscure enough that they would be impossible to predict through logical inference. We provided ~100 word passages around 60 such noun phrases and asked four models to fill in the blanks (Qwen-2.5-72B, R1-distill-Llama-70B, GPT-4.1, GPT-5.4-medium).

None of the models answered any questions correctly: the score was zero out of 240 tries. To look more carefully for a memorization effect, we reframed the test as multiple choice, and measured Pearson correlation between a model's average skill score guessing proper names in a book and average skill scores on benchmark questions based on the same book. We found no evidence of a relationship ($n = 64, r=-0.014, p=0.911$). This weak effect makes sense given \citet{chang2023speak}'s finding that the risk of memorization is highest for books frequently quoted on the web. We explicitly avoided books with that level of fame. 

To preserve the innocence of future models, we share a sample of questions online rather than the full list. But this benchmark is mostly intended to drive research, not commercial competition. The past is finite, and it is now too late to create a genuinely sequestered test set. Historical benchmarks should allow researchers to exclude specified volumes from training, not try to create a test that can repel concerted attempts at memorization.

\section{Benchmark structure}

Our goal, as explained above, was not to invent a new ontology for the representation of culture, but to transpose existing methods of evaluating language models to the period 1831-1930. We therefore freely borrowed question categories from existing benchmarks. Our inference questions could be compared to ARC, for instance, and knowledge questions to MMLU \citep{AI2reasoning2018, MMLU2020}. We were constrained only by the requirement that ground truth answers should be available verbatim in some period text. That is why we reframed summarization tasks as ``topic sentence'' questions, asking the model to supply a (masked) introductory sentence for a paragraph.

\begin{table}[t]
\centering
\small
\begin{tabular}{lrr}
\toprule
Reasoning type & $N$ & \% \\
\midrule

\multicolumn{3}{l}{\textit{Cloze tasks}} \\
structured cloze -- sentence & 160 & 18.5 \\
structured cloze -- phrase & 155 & 17.9 \\
topic sentence & 44 & 5.1 \\

\addlinespace
\multicolumn{3}{l}{\textit{Generation tasks}} \\
constrained generation & 158 & 18.2 \\
character modeling & 115 & 13.3 \\

\addlinespace
\multicolumn{3}{l}{\textit{Knowledge and inference}} \\
knowledge & 94 & 10.9 \\
inference & 78 & 9.0 \\
abstention & 62 & 7.2 \\

\midrule
Total & 866 & 100.0 \\
\bottomrule
\end{tabular}
\caption{Distribution of reasoning types in Chronologic-EN-1.0.}
\label{tab:reasoning-types}
\end{table}

\subsection{Types of questions} 

Our prototype benchmark has 866 questions in total. Each is assigned one of the following ``reasoning types'' that describes the task. For examples of each type, see Appendix \ref{sec:appendix-questions}.

\textbf{Structured cloze.} Contemporary benchmarks de-emphasize cloze tasks, since they are implicitly covered by predictive loss. But we cannot assume that all models will have been trained on period text, so cloze remains an important foundation here. Questions are structured to reward logical inference, and typically require infill rather than simple continuation.

\textbf{Topic sentence.} Here the question takes the form of a paragraph with a missing topic sentence. This tests the model's ability to grasp the main idea of a passage and summarize it like a writer in a specified period and genre. 

\textbf{Constrained generation.} The question specifies a subject or form, while the metadata frame specifies a social context. The model has to write a short passage of prose or verse that fits both constraints. For instance an open-ended question (``What can be affirmed about the fate of the soul after death?'') is constrained by the instruction to provide an answer that might have appeared in \emph{The Theosophical Forum} in 1903. Sometimes the question may include arbitrary formal requirements (e.g., exactly two sentences) modeled on the constraints in IFEval \citep{zhou2023instructionfollowing}.

\textbf{Character modeling.} Each question provides descriptions of a character drawn from a fictional work and then requests a line of dialogue the character might speak in a briefly summarized situation. This tests a model's understanding of genre, and also of period assumptions about character and manners (how do young people speak to parents, for instance?)

\textbf{Knowledge.} These test a model's ability to provide facts or define terms. Many answers are drawn from period reference books. 

\textbf{Inference.} Based largely on period textbooks, these questions test a model's ability to draw conclusions from information provided. In addition to questions about mathematics, grammar, rhetoric, and engineering, we provide passages of verse and ask the model to describe the poetic form exemplified. 

\textbf{Abstention.} Test whether models can express appropriate uncertainty and decline questions that require information not known at the time.

\begin{table*}[ht]
    \centering
    \includegraphics[width=\textwidth]{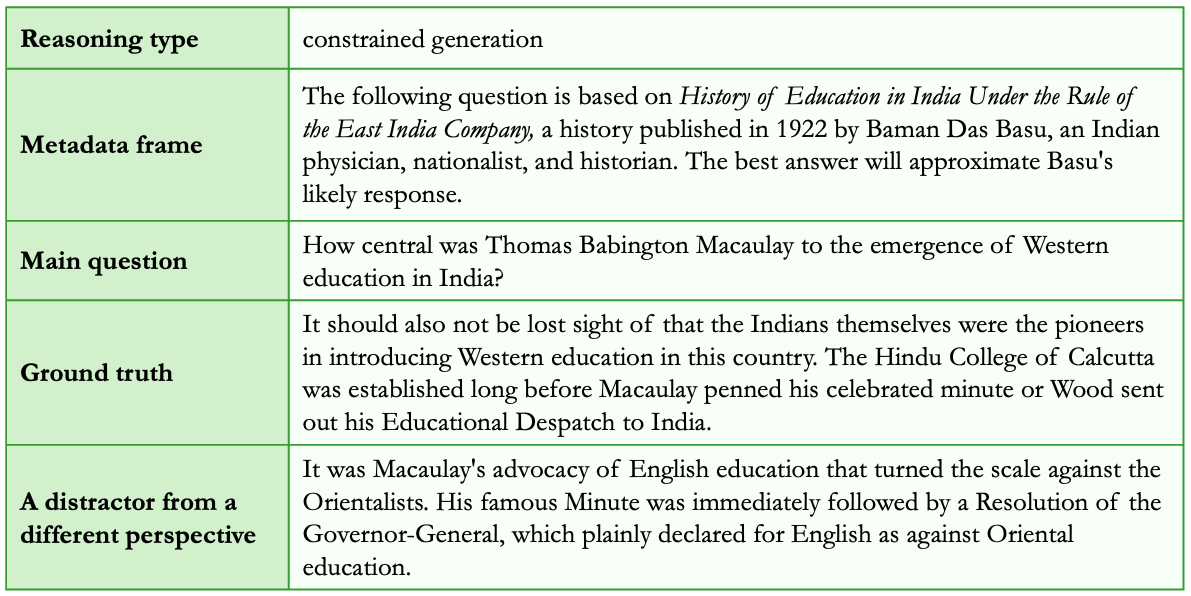}
    \caption{An example of the relationship between metadata frame and main question in a ``book context'' question. This sample distractor comes from \emph{Report of the Indian Education Commission} (Hunter Commission), issued by the Government of India under British colonial rule, 1882. Other distractors came from 21c sources.}
    \label{tab:samplequestion}
\end{table*}

\subsection{Context dependence} 

If you ask a model for a ``Victorian'' perspective on a question, stilted references to propriety will ensue. But people in 1876 were actually separated by as many differences as people in 2026. To test a model's recognition of differences within a period, every question in our benchmark is accompanied by a ``metadata frame'' that specifies a context.

Of course, some questions (``What is the capital of France?'') were not topics of fierce disagreement. So the frames for most knowledge and inference questions are sketched loosely, specifying only a date and nationality (``an American encyclopedia published in 1897''), and asking for an answer that reflects knowledge available at the time. We describe this as ``world context.'' 

Other questions depend heavily on a particular social perspective. Here we typically provide ``book context'': a phrase that characterizes the genre of the book and the author's background or occupation, along with date of first publication.

\subsection{Admissible variation}

Although our ground truth answers are drawn from historical texts, we understand the texts as samples from a larger population. The goal is not for models to precisely match these answers, but to provide answers that fall within a penumbra of contextually plausible variation around them.

One way we establish that penumbra is to provide multiple ground truth answers for 10\% of questions. In some cases these are different passages from a single book; in other cases the question asks for the perspective of a group, and answers come from different members of the group. For instance, what might a liberal British politician say about the ``necessaries of life'' in the early 1840s? Having multiple ground truths helps us calibrate the range of admissible variation, especially in free-generation scoring.

\section{Experiments}

The benchmark can be evaluated in two different ways. 

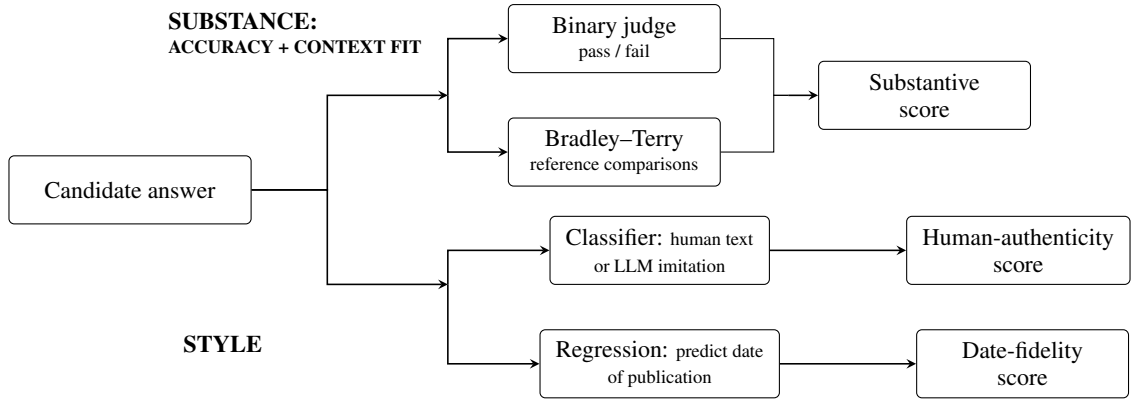
\begin{figure*}[th!]
\centering

\begin{tikzpicture}[
    x=1cm,
    y=1cm,
    box/.style={
        draw,
        rounded corners=2pt,
        align=center,
        minimum height=9mm,
        minimum width=28mm,
        inner xsep=6pt,
        font=\small
    },
    input/.style={
        box,
        minimum width=32mm
    },
    arrow/.style={
        ->,
        >=stealth,
        semithick
    },
    lane/.style={
    font=\small\bfseries,
    align=left,
    inner sep=0pt
    },
    forklabel/.style={
        font=\scriptsize\bfseries
    },
    note/.style={
        font=\scriptsize,
        align=center
    }
]


\node[input] (input) at (0,0)
    {Candidate answer};

\node[lane, anchor=south west] at (0.5,1.8)
    {SUBSTANCE:\\[-1pt]
     {\scriptsize ACCURACY + CONTEXT FIT}};

\node[lane, anchor=south west] at (0.7,-2.15)
    {STYLE};


\coordinate (subfork) at (4.2,1.25);

\node[box] (binary) at (6.4,2.0)
    {Binary judge\\[-1pt]
     {\scriptsize pass / fail}};

\node[box] (bt) at (6.4,0.5)
    {Bradley--Terry\\[-1pt]
     {\scriptsize reference comparisons}};

\node[box] (subscore) at (10.5,1.25)
    {Substantive\\score};

\draw[arrow]
    (input.east) -- ++(1.0,0) |- (subfork);

\draw[arrow]
    (subfork) |- (binary.west);

\draw[arrow]
    (subfork) |- (bt.west);


\coordinate (submerge) at (8.7,1.25);

\draw
    (binary.east) -- ++(0.7,0) |- (submerge);

\draw
    (bt.east) -- ++(0.7,0) |- (submerge);

\draw[arrow]
    (submerge) -- (subscore.west);


\coordinate (stylefork) at (4.2,-1.25);

\node[box] (auth) at (7.0,-0.8)
    {Classifier: \scriptsize human text \\ \scriptsize or LLM imitation};

\node[box] (date) at (7.0,-2.3)
    {Regression: \scriptsize predict date \\ \scriptsize of publication};

\draw[arrow]
    (input.east) -- ++(1.0,0) |- (stylefork);

\draw[arrow]
    (stylefork) |- (auth.west);

\draw[arrow]
    (stylefork) |- (date.west);


\node[box, right=18mm of auth] (authscore)
    {Human-authenticity\\score};

\node[box, right=18mm of date] (datescore)
    {Date-fidelity\\score};

\draw[arrow] (auth.east) -- (authscore.west);
\draw[arrow] (date.east) -- (datescore.west);

\end{tikzpicture}

\caption{
Scoring paths for a freely-generated answer.
Substantive scoring uses either binary pass/fail evaluation (for 520 questions) or Bradley--Terry partial-credit scoring (for 346); the two paths produce a common substantive score, which can be aggregated by question type. Style evaluation applies both authenticity and
date-fidelity measures to all answers of sufficient length.
}
\label{fig:scoring-pipeline}
\end{figure*}

\subsection{Likelihood scoring}

With open-weight models we can directly measure the probability of an appropriate response. The per-token log-likelihoods of one or more ground truth answer(s) can be compared to the probabilities of distractors that are either factually wrong or stylistically anachronistic. Results can be assessed either through Brier scores or by interpreting the highest-probability completion as the model's ``choice.''

\subsection{Free generation}
\label{model-as-judge}
With closed models and reasoning models, it can be hard to measure likelihood in a meaningful way. A more flexible way to measure performance is to allow models to answer questions freely and then evaluate the answers. But judging the fit between an arbitrary answer and a given historical context is a challenging task.

\subsubsection{Style judgment}

To be historically plausible, an answer must be expressed in language that fits the specified period. Neither human judges nor modern LLMs are trained to make this kind of stylistic judgment, so we train two DeBERTa models to discern different dimensions of stylistic fit:

\begin{itemize}
\item A classifier is trained to distinguish authentic historical text from LLM-generated paraphrase, infill, or continuation.
\item A 36-way ordinal softmax over date bins is trained against soft Gaussian targets to predict publication date (as a distribution), and measure divergence from the actual date targeted by a question.
\end{itemize}

Rather than trusting either judge's raw output, we convert it to a percentile among date- and length-matched authentic passages scored by the same model, so that a text indistinguishable from period prose yields uniform percentiles. The reported 0–100 scores measure divergence from that uniform ideal, rescaled against the divergence authentic passages themselves produce at the same sample size. (See Appendix \ref{sec:appendix-scoring}.)

\subsubsection{Substantive judgment}

Judging the substantive aptness of an answer is equally challenging. Answers should follow instructions, and should fit the social context described either factually (when the question provides ``world context'') or rhetorically (when it provides ``book context''). In previous work we found that human readers are able to distinguish authentic historical passages from LLM imitations when given head-to-head comparisons---but not necessarily able to rate passages presented in isolation \cite{underwood_nelson_wilkens_2025_anachronism}. We make use of that insight here by staging judgment as a series of head-to-head comparisons.

\textbf{Pass/fail scoring.} For relatively simple questions, we make a single comparison between a model's response and an (unlabeled) ground truth answer. If the judge (Claude Sonnet 4.6) perceives the model's response as better than or equal to ground truth, it passes the test. We assess accuracy by presenting the judge with pairs of unlabeled answers where both are known to be correct, or one is known to be incorrect. Pass/fail scoring is only used for questions where the LLM judge is generally successful at this task; average Type I and II error rates are both below 5\%.

\textbf{Partial credit (Bradley-Terry) scoring.} Binary scoring works well for knowledge and inference questions, which often have a single right answer. It also works for most cloze questions, where the substantive part of judgment is mainly to confirm logical coherence. But in constrained generation tasks, the substantive correctness of an answer depends on nuanced judgments about fit with a specific social context. Is this really something a Quaker periodical in the U.S. could have said about treatment of Indians in the 1860s?

The questions of degree involved here would be difficult to calibrate with a single judgment. But we have 3-7 distractors for all questions, and multiple ground truths for many of our most difficult questions. We can therefore compare each candidate to multiple references, in order to place it on a scale. We create the scale for each question by making forced-choice comparisons between all pairs of distractors and ground truth answers, and use a Bradley-Terry model to infer latent strengths for each reference, much as chess players receive Elo scores from a series of matches. We have not found an exact precedent for this item-level latent quality scale, but pairwise BT evaluation of language models is generally well established \cite{li2025arenahard, lin2026jptlbenchanchoredpairwisellm, mozafari2026qdaps}. As noted in \citet{don-yehiya2026mediocrity}, discriminative strength depends on anchors that measure shades of gray in the middle of the scale---for examples, see Appendix \ref{sec:appendix-bt-example}.

In order to guide the judge (Claude Sonnet 5) in making these forced-choice comparisons, a rubric is developed for each question, which takes the form of a list of reasons an answer might be rejected. Whenever a ground truth answer is not involved in the current comparison, the judge is also shown that ground truth answer as a guide.

\begin{figure}[t]
    \centering
    \includegraphics[width=\linewidth]{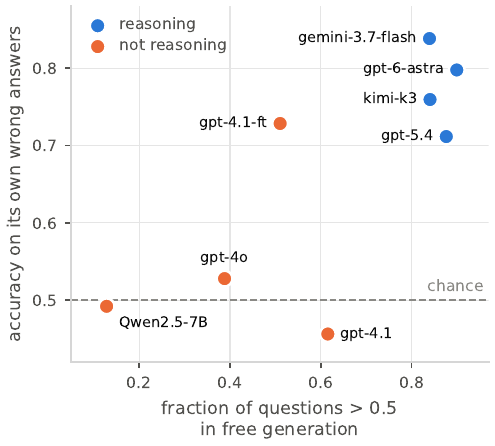}
    \caption{Initial generative success rates are on the x axis; discriminative accuracy in rejecting the model's own $< 0.5$ answers in MCQ format is on the y axis.}
    \label{figure:self-rejection}
\end{figure}

To turn Bradley-Terry win probabilities into calibrated benchmark scores, we have to measure the penumbra of acceptable variation around ground truth. This is possible because $80$ questions have multiple ground truth answers, allowing us to hold out one ground truth answer and score it like a candidate. A logistic model is then trained to map the Bradley-Terry strength of a held-out answer to its probability of being acceptable. We pin the intercept so the average held-out ground truth answer scores $0.90$. For details of the calibration process, see Appendix \ref{sec:appendix-scoring}. We also asked six human judges to rank pairs of answers in order to validate this ranking. See Appendix \ref{sec:appendix-human} for details.

\subsection{Why multiple choice is not used}

We are interested in a model's ability to generate historically plausible responses. Multiple choice works as a proxy for this generative task only if it is roughly as difficult to recognize a correct answer as to produce one. 

\begin{table}[t]
\centering
\small
\setlength{\tabcolsep}{3pt}
\begin{tabular}{lcccc}
\toprule
\textbf{Model} & \textbf{Brier} & \textbf{Cloze} & \textbf{Gen.} & \textbf{Know.} \\
\midrule
Talkie 1930 13B base       & \textbf{0.1574} & \textbf{39.5} & \textbf{24.9} & \textbf{44.0} \\
Talkie 1930 13B it         & 0.1593 & 37.4 & 23.8 & 25.1 \\
Qwen 2.5 7B finetune             & 0.1601 & 36.2 & 14.9 & 34.4 \\
Qwen 2.5 72B               & 0.1607 & 10.8 &  8.8 & 38.0 \\
Qwen 3.5 35B-A3B base  & 0.1623 & 12.6 &  8.0 & 37.1 \\
Talkie web 13B base        & 0.1627 & 18.2 & 13.6 & 34.1 \\
Qwen 2.5 7B                & 0.1628 &  8.6 &  9.5 & 37.0 \\
\bottomrule
\end{tabular}
\caption{Open-weight models scored by per-token likelihoods assigned to answers. Overall Brier score (lower is better) and per-category accuracy (\%, higher is better) on cloze, constrained generation, and knowledge/inference questions.}
\label{tab:leaderboard}
\end{table}

But discriminative framing is known to be easier for human test-takers, and in some cases for language models \citep{chandak2025answer, mee2024mcq_saq}. To measure the scope of the problem for our task, we took freely-generated model answers that our scoring pipeline had scored less than $0.5$, and offered those answers as distractor options to the same model that originally wrote them, in head-to-head forced choices with ground truth. All reasoning models were able to reject their own wrong answers in a multiple-choice setting. Accuracy ranged from 71\% to 84\%, with $p << .001$. 

This result is interestingly at odds with 2024-era scholarship, which reports that ``language models seem to acquire generation abilities more effectively than understanding'' \citep{west2024generative}. It is possible that historical simulation is an unusual task, much easier to critique than to produce. It is also possible, as Figure \ref{figure:self-rejection} hints, that reasoning models really changed the balance between generative and discriminative power.
\begin{table*}[ht!]
  \centering
  \small
  \begin{tabular}{llccccccc}
  \toprule
  \textbf{Model} & \textbf{Effort} & \textbf{Cloze} & \textbf{Generation} & \textbf{Knowledge} & \multicolumn{1}{c}{\thead[c]{\textbf{Date}\\ \textbf{fidelity}}} & \multicolumn{1}{c}{\thead[c]{\textbf{Error}\\ \textbf{in years}}} & \multicolumn{1}{c}{\thead[c]{\textbf{Human}\\ \textbf{authenticity}}} \\
  \midrule
  Talkie 1930 13B base &    & 1.5  & 12.6 & 20.3 & 54.4  & 30.2  & 32.6 \\
  Talkie 1930 13B it   &    & 6.3  & 32.9 & 22.6 & 87.7  & 12.6  & 32.1 \\
  \addlinespace[4pt]
  Qwen2.5 7B          &    & 7.3  & 13.2 & 25.5 & 4.8   & 105.2 & 3.7 \\
  Qwen2.5 72B it       &    & 19.1 & 20.2 & 50.2 & 13.8  & 84.2  & 5.4  \\
  \addlinespace[4pt]
  Gemini 3.7 flash     & high   & 85.8 & 60.9 & \textbf{98.2} & 76.7  & 14.9  & 27.4 \\
  \addlinespace[4pt]
  Kimi K3              & max    & 85.7 & 66.6 & 94.4 & 94.8  & 6.6   & 24.0 \\
  \addlinespace[4pt]
  GPT-4o 2024-08-06              &    & 35.0 & 26.0 & 65.0 & 19.6  & 79.1  & 5.7 \\
  GPT-4.1              &    & 65.6 & 44.0 & 68.9 & 65.8  & 27.5  & 11.9 \\
  GPT-4.1 finetune     &    & 38.9 & 53.8
  & 62.7 & 98.9 & 2.9   & \textbf{88.4} \\
  GPT-5.4              & medium & 89.9 & 68.0 & 93.1 & 94.2  & \textbf{-1.9}  & 28.1 \\
  GPT-5.4 wrong contexts & medium & 83.7 & 54.3 & 85.9 & 91.0  & -3.9  & 26.9 \\
  GPT-5.6              & high   & 91.0 & \textbf{71.7} & 92.3 & 92.3  & 6.6   & 28.7 \\
  GPT-6-Astra              & high & \textbf{93.0} & 71.4 & 96.1 & \textbf{99.3}  & 4.0   & 23.7 \\
  \bottomrule
  \end{tabular}
  \caption{Free-generation results (higher is better for all columns except error in years, a signed quantity where lower absolute magnitude is better). For scoring method, see Figure \ref{fig:scoring-pipeline}; substantive score is broken down here by the reasoning type of questions. For confidence intervals, see Appendix \ref{sec:appendix-confidence}. For fine-tuning of GPT-4.1, Appendix \ref{sec:appendix-tuning}.}
  \label{tab:free-gen-results}
\end{table*}
  
\section{Results}

\subsection{Likelihood scores}

The results of likelihood scoring appear in Table \ref{tab:leaderboard}. Models are sorted by overall Brier score, which is in principle the most meaningful measure here, since it assesses not only top-1 accuracy but calibration. The Talkie-1930 base model dominates all categories, as one might expect: it was pretrained on text from the nineteenth and twentieth centuries and is well suited to this task. Models trained primarily on late 20c and 21c text can perform below random accuracy on cloze tasks and constrained generation tasks, because the answers for these questions are long enough for style to determine their likelihood in the model, and modern answers may seem more likely than authentic ones. Finetuning on period text significantly improves the performance of the otherwise weak Qwen 2.5 7B. (For specific question types grouped under the three columns see Table \ref{tab:reasoning-types}.)

\subsection{Free generation scores}
Table \ref{tab:free-gen-results} presents a very different picture. Recently-released reasoning models dominate substantive aspects of the benchmark. They achieve > 90\% in cloze and knowledge tasks, but continue to struggle with generative tasks (e.g. writing passages that might have appeared in a given context). Scores on this aspect peak at 71.7\%.

Talkie-1930 is not a particularly strong candidate when freely generated answers are scored. The instruction-tuned version performs best---but the model's tuning appears to struggle with the complexity of even cloze (infill) questions. On generative tasks, however, it does outperform other open-weight models and even GPT-4o.

More interestingly, Talkie-1930 is only modestly strong where it might be expected to excel: in targeting the style of a specified decade. Its date fidelity is better than that of Qwen2.5, Gemini 3.7, and an untuned GPT-4.1, but not equal to GPT 5.x models or finetuned 4.1. Here it is worth recalling that training Talkie only on text written before 1930 doesn't guarantee that it will understand differences within its corpus. Moreover, while Talkie's training data covers the whole century 1831-1930, it is likely to be larger toward the end, since book publication increases. The model's tendency to sound more modern than a randomly selected passage within the 19c is not surprising.

The other stylistic score, human authenticity, shows only that no model fully escapes a shared LLM style. The GPT-4.1 finetune is a partial exception, but fine-tuning has harmed some of its substantive scores while making style less typical.

\subsection{Metadata frames do usefully guide models}

The ``wrong contexts'' row in Table \ref{tab:free-gen-results} lists the results of an experiment where GPT-5.4 was given the benchmark with permuted metadata frames. The question itself remained correct, but the model was encouraged to respond within an inappropriate sociohistorical context. Scores drop by 6-14 points across the three question categories relative to the same model's performance with correct metadata frames. The decline is especially marked for the ``generation'' category, which includes many questions where social context is the primary factor determining answer appropriateness.

\section{Discussion}

Since large commercial reasoning models performed well on the free-generation scoring path, it may appear at first glance that special-purpose historical models are not needed. But we would encourage a more cautious interpretation of Tables \ref{tab:leaderboard} and \ref{tab:free-gen-results}. 

The two measurement strategies we deployed measure different things. Direct measurement of likelihood can show that a model actually reflects the probabilities that governed a period's writing. A model that scored high here would be more than a chatbot: it could be used to probe the relation between language and context. To study the transformation of manners, for instance, we could take advice from a conduct book in 1910 and vary metadata frames to ask how the probability of each statement varies across time, nationality, or genre (perhaps some things officially disreputable were already the norm in fiction).  No model we tested achieved a score that provides a firm foundation for this experiment yet, but it is the sort of experiment this part of the benchmark could validate.

Success in free generation demonstrates something different. It shows that a model can---perhaps after a lengthy chain of thought---produce at least \emph{one} contextually appropriate response to a question. This might be adequate if the goal is to create a chatbot plausible enough for student interaction, but it doesn't provide the flexible foundation for research an open-weight model would provide. (For one thing, free generation doesn't demonstrate that you can show the model a passage and get back a well-calibrated probability.)

Our benchmark suggests there is still some distance to cross before models achieve even this more accessible goal. Cloze and knowledge questions are close to saturated---unsurprisingly, since those are fairly easy tasks for reasoning models. But generation is the part of the benchmark most immediately relevant to a model's simulation of the past, and performance there tops out at 71.7. This isn't terrible, especially since partial-credit questions are scaled with mean ground truth performance at 90.0. But it means chatbots asked to simulate the past produce output that is perceptible as pastiche. 

We see no evidence, however, that this obstacle will prove insuperable. The last two years have seen significant progress on generation quality. Scores increase 18 points from GPT-4o (Summer 2024) to GPT-4.1 (Spring 2025), and another 28 points to GPT-5.6 (Summer 2026). It seems unlikely this task was a commercial priority, so perhaps the improvement was a side-effect of reasoning. Indeed, we find that strong reasoning models are not only able to detect flaws in their own responses, but often able to provide a rationale for their judgment and identify tell-tale signs of pastiche. (See Appendix \ref{sec:appendix-diagnostic}.) It seems possible that the first plausible historical chatbots will not be produced by blinding models to ensure naïveté, but by fostering higher levels of self-conscious reflection.

Perhaps we should have expected that historical knowledge would have value for this task. Hans-Georg Gadamer and others have argued that historical understanding works not by reconstructing an uncontaminated past, but by creating a ``fusion of horizons'' where different worlds meet to reflect on their differences \citep{Gadamer2004-GADTAM}. This is true in any case about historical models trained to interact with contemporary interlocutors: they will have to lose some of their purity, in order to become the kind of thing that can talk to us.

\subsection{Future work}

If likelihood scoring and free generation test fundamentally different applications, it may not make sense to pit purpose-built models like Talkie-1930 against frontier reasoning models. They excel at different tasks. 

But both kinds of models could be improved, and this benchmark aims to guide both projects. For open-weight models, cloze and knowledge/inference questions remain important tests. Date fidelity is also still a challenge: as Benjamin Breen has shown, pretraining a model on nineteenth-century text doesn't necessarily give it a coherent historical vantage point, let alone an ability to shift between vantage points \citep{breen2026vintage}.

Recent frontier models have a remarkable grasp of period style: if GPT-6, for instance, is told to write something from 1884, a discriminative judge usually places the response within four years of that target. But inspection of answers shows that they are still distinguished from ground truth by a tendency to overplay and caricature the aspects of historical prose that depart from contemporary social norms. For instance, dialogue in nineteenth-century novels often acquires a ``Wardour Street'' coloration, with \emph{thee}s and \emph{thou}s inappropriate for the context. Female characters may become exaggeratedly pious. But since reasoning models can often diagnose the flaws in their own responses, there is reason to suspect that these weaknesses can be addressed.

The likelihood-scored part of this benchmark has plenty of headroom. As models become stronger, we will try to keep the free-generation part competitive as well by designing questions with longer answers and tougher distractors. Expansion to other periods and languages is desirable, but will require considerable labor and human expertise if benchmarks intend to challenge frontier models. The discriminative strengths of reasoning models may need to be harnessed for question design.

Our benchmark, and code to administer it, are accessible at \url{https://github.com/Historical-AI-Lab/Chronologic-EN}. 

\section{Limitations}

This prototype only covers one century of writing in one language, and is only based on 187 sources. Although we strove to maximize diversity, 866 questions cannot claim to completely cover written culture in this period. Questions are not distributed uniformly; only about a third cover the first 50 years, while two-thirds cover 1881-1930. This is arguably congruent with the distribution of sources available for training models.

Grounding a benchmark in published writing means grounding it disproportionately in elite experience. A model that claimed to represent the mentality of a population sample drawn randomly from the streets of 1850 would need to be tested against a different sort of benchmark, very difficult indeed to construct.

Some distractors in the benchmark are created by contemporary language models. In particular, many distractors were created by gpt-oss:20b and mistral-small:24b. These models might perform poorly on likelihood-scored versions of the test because the wrong answers are particularly tempting for them. 

Contributing distractors creates less of a bias in free generation. We do include distractors generated by some of the models taking the test, and if models repeat themselves, it is possible that this will lead to effective ties in a pairwise comparison with their own distractor. But anchor strength is determined by relationships between a whole system of answers, so providing a distractor is not necessarily a disadvantage for a model. If the distractor was strong, their similar answer will also end up with a strong score. (Indeed, it is theoretically possible for a strong distractor to exceed a ground truth answer.) 

The bias that does become significant for free generation is that we have relied on Claude-family models as judges. To evaluate products from Anthropic, we would need to validate an alternate judge.

The style judgments reported in Table \ref{tab:free-gen-results} are not based on precisely the same sets of questions in every case. Because answers of less than 6 words are excluded from authenticity measurement, and those less than a sentence long are excluded from date prediction, slightly different sets of questions may be excluded, depending on model verbosity.

The Bradley-Terry models used in partial-credit scoring produce primarily an ordinal ranking of answers. The distances between ranks are not always well calibrated, because it is common for the graph of pairwise comparisons to divide into two subgraphs, where all members of A (e.g. ground truths) sweep all members of B (e.g. distractors). In such a case, the gap between A and B will be shaped primarily by a scaling prior. 

For this reason we don't use raw latent strengths from the BT process, but calibrate them through a second model (see Appendix \ref{sec:appendix-scoring}). These scores have been directionally validated by human judgment: we can report that every 10 points of difference between the ``generation'' scores of two models in Table \ref{tab:free-gen-results} corresponds to an additional 5\% probability (above a coin flip) that a human reader will prefer the higher-scoring model's response to a random question. Once the gap between models reaches 11 points, the 95\% confidence interval on this preference excludes 0.5, so gaps of that size should be treated as having significance $p < .05$.

Calibration of partial-credit scores is validated mainly by the variation between ground truths; the average score of a held-out ground truth answer is pinned at 0.9. This means that some portion of the difference between ``knowledge'' and ``generation'' scores in Table \ref{tab:free-gen-results} may be attributable to the partial-credit scoring of generation questions, which creates a likely ceiling of 0.9 for that column rather than 1.0. In calibrating scores, we do also provide examples of partly-acceptable model-generated answers, to avoid creating a scale where nothing short of ground truth can approach 0.9.

Human performance on this benchmark has not been measured. If it is measured, we advise interpreting the result cautiously, because this is not a task human beings are normally expected to perform. In asking language models to represent past writing practices, we are not trying to replicate human performance---but trying to measure a system that does something humans ordinarily cannot.

\section*{Acknowledgments}

Work on this project was supported by the Humanities and AI Virtual Institute (HAVI), funded by Schmidt Sciences. Early planning stages were also supported by UIUC Research Board grant RB25120.

Computational resources were provided by DELTA at the National Center for Supercomputing Applications (NCSA) through allocation HUM240002 from the Advanced Cyberinfrastructure Coordination Ecosystem: Services and Support (ACCESS) program \cite{boerner2023access}, which is supported by U.S. National Science Foundation grants \#2138259, \#2138286, \#2138307, \#2137603, and \#2138296.

\bibliography{custom}

@article{rao2024normad,
  title={Norm{A}d: A Framework for Measuring the Cultural Adaptability of Large Language Models},
  author={Rao, Abhinav and Yerukola, Akhila and Shah, Vishwa and Reinecke, Katharina and Sap, Maarten},
  journal={arXiv preprint arXiv:2404.12464},
  year={2024},
  doi={10.48550/arXiv.2404.12464},
  url={https://arxiv.org/abs/2404.12464}
}

@article{shi2024culturebank,
  title={Culture{B}ank: An Online Community-Driven Knowledge Base Towards Culturally Aware Language Technologies},
  author={Shi, Weiyan and Li, Ryan and Zhang, Yutong and Ziems, Caleb and Yu, Chunhua and Horesh, Raya and Abreu de Paula, Rog{\'e}rio and Yang, Diyi},
  journal={arXiv preprint arXiv:2404.15238},
  year={2024},
  doi={10.48550/arXiv.2404.15238},
  url={https://arxiv.org/abs/2404.15238}
}

@article{adilazuarda2024culture,
  title={Towards Measuring and Modeling ``Culture'' in {LLMs}: A Survey},
  author={Adilazuarda, Muhammad Farid and Mukherjee, Sagnik and Lavania, Pradhyumna and Singh, Siddhant and Aji, Alham Fikri and O'Neill, Jacki and Modi, Ashutosh and Choudhury, Monojit},
  journal={arXiv preprint arXiv:2403.15412},
  year={2024},
  doi={10.48550/arXiv.2403.15412},
  url={https://arxiv.org/abs/2403.15412}
}

@inproceedings{zhou2025culture,
    title = "Culture is Not Trivia: Sociocultural Theory for Cultural {NLP}",
    author = "Zhou, Naitian  and
      Bamman, David  and
      Bleaman, Isaac L.",
    editor = "Che, Wanxiang  and
      Nabende, Joyce  and
      Shutova, Ekaterina  and
      Pilehvar, Mohammad Taher",
    booktitle = "Proceedings of the 63rd Annual Meeting of the Association for Computational Linguistics (Volume 1: Long Papers)",
    month = jul,
    year = "2025",
    address = "Vienna, Austria",
    publisher = "Association for Computational Linguistics",
    url = "https://aclanthology.org/2025.acl-long.1256/",
    doi = "10.18653/v1/2025.acl-long.1256",
    pages = "25869--25886",
    ISBN = "979-8-89176-251-0"
}

@article{varnum2024historical_llms,
  author  = {Varnum, Michael E. W. and Baumard, Nicolas and Atari, Mohammad and Gray, Kurt},
  title   = {Large language models based on historical text could offer informative tools for behavioral science},
  journal = {Proceedings of the National Academy of Sciences},
  year    = {2024},
  volume  = {121},
  number  = {42},
  pages   = {e2407639121},
  doi     = {10.1073/pnas.2407639121}
}

@misc{bisbee2023synthetic,
  author = {Bisbee, James and Clinton, Joshua D. and Dorff, Cassy and Kenkel, Brenton and Larson, Jennifer M.},
  title = {Synthetic Replacements for Human Survey Data? {T}he Perils of Large Language Models},
  institution = {SocArXiv},
  year = {2023},
  month = {5},
  note = {{SocArXiv} preprint},
  doi = {10.31235/osf.io/5ecfa}
}

@misc{boelaert2024machine,
  author = {Boelaert, Julien and Coavoux, Samuel and Ollion, Étienne and Petev, Ivaylo D. and Präg, Patrick},
  title = {Machine Bias. {G}enerative Large Language Models Have a Worldview of Their Own},
  institution = {SocArXiv},
  year = {2024},
  month = {4},
  note = {{SocArXiv} preprint},
  doi = {10.31235/osf.io/r2pnb}
}

@article{underwood_nelson_wilkens_2025_anachronism,
  author  = {Underwood, Ted and Nelson, Laura K. and Wilkens, Matthew},
  title   = {Can Language Models Represent the Past without Anachronism?},
  journal = {arXiv preprint arXiv:2505.00030},
  year    = {2025},
  doi     = {10.48550/arXiv.2505.00030}
}

@misc{goettlich_et_al_2025_history_llms_blog,
  author       = {Göttlich, Daniel and Loibner, Dominik and Voth, Hans-Joachim},
  title        = {History {LLMs}: Giving the Past a Voice with Large Language Models},
  year         = {2025},
  month = {December},
  howpublished = {Broadstreet},
  url = {https://www.broadstreet.blog/p/history-llms-giving-the-past-a-voice}
}

@article{zhang_et_al_2025_culturescope,
  author  = {Zhang, Jinghao and Jiang, Sihang and Guo, Shiwei and Chen, Shisong and Xiao, Yanghua and Feng, Hongwei and Liang, Jiaqing and He, Minggui and Tao, Shimin and Ma, Hongxia},
  title   = {Culture{S}cope: A Dimensional Lens for Probing Cultural Understanding in {LLMs}},
  journal = {arXiv preprint arXiv:2509.16188},
  year    = {2025},
  doi     = {10.48550/arXiv.2509.16188}
}

@article{foster2018culture_computation,
  author  = {Foster, Jacob G.},
  title   = {Culture and Computation: Steps to a Probably Approximately Correct Theory of Culture},
  journal = {Poetics},
  year    = {2018},
  volume  = {68},
  pages   = {144--154},
  doi     = {10.1016/j.poetic.2018.04.007}
}

@article{chandak2025answer,
  title={Answer Matching Outperforms Multiple Choice for Language Model Evaluation},
  author={Chandak, Nikhil and Goel, Shashwat and Prabhu, Ameya and Hardt, Moritz and Geiping, Jonas},
  journal={arXiv preprint arXiv:2507.02856},
  year={2025},
  url={https://arxiv.org/abs/2507.02856}
}

@article{mee2024mcq_saq,
  author  = {Mee, Janet and Pandian, Ravi and Wolczynski, Justin and Morales, Amy and Paniagua, Miguel and Harik, Polina and Baldwin, Peter and Clauser, Brian E.},
  title   = {An experimental comparison of multiple-choice and short-answer questions on a high-stakes test for medical students},
  journal = {Advances in Health Sciences Education},
  year    = {2024},
  volume  = {29},
  number  = {3},
  pages   = {783--801},
  doi     = {10.1007/s10459-023-10266-3}
}

@book{Gadamer2004-GADTAM,
	address = {New York},
	author = {Hans{-}Georg Gadamer},
	publisher = {Continuum},
	title = {Truth and Method},
	year = {2004}
}

@inproceedings{chang2023speak,
  title     = {Speak, Memory: An Archaeology of Books Known to {ChatGPT}/{GPT}-4},
  author    = {Chang, Kent K. and Cramer, Mackenzie and Soni, Sandeep and Bamman, David},
  booktitle = {Proceedings of the 2023 Conference on Empirical Methods in Natural Language Processing},
  year      = {2023},
  address   = {Singapore},
  publisher = {Association for Computational Linguistics},
  pages     = {7312--7327},
  doi       = {10.18653/v1/2023.emnlp-main.453},
  url       = {https://aclanthology.org/2023.emnlp-main.453/}
}

@misc{idi_institutional_books_1_2024,
  author       = {{Institutional Data Initiative}},
  title        = {{Institutional Books 1.0}},
  year         = {2025},
  howpublished = {\url{https://huggingface.co/datasets/institutional/institutional-books-hl}},
  note         = {Harvard University},
}

@misc{poibeau2026historical,
  author = {Poibeau, Thierry},
  title  = {What Do Historical Language Models Model?},
  year   = {2026},
  url    = {https://hal.science/hal-05662234v1},
  note   = {{HAL} preprint, hal-05662234, version 1}
}

@misc{cargnelutti2025institutionalbooks10242b,
      title={Institutional {B}ooks 1.0: A 242{B} token dataset from {H}arvard {L}ibrary's collections, refined for accuracy and usability}, 
      author={Matteo Cargnelutti and Catherine Brobston and John Hess and Jack Cushman and Kristi Mukk and Aristana Scourtas and Kyle Courtney and Greg Leppert and Amanda Watson and Martha Whitehead and Jonathan Zittrain},
      year={2025},
      eprint={2506.08300},
      archivePrefix={arXiv},
      primaryClass={cs.CL},
      url={https://arxiv.org/abs/2506.08300}, 
}

@misc{levine2026talkie,
  title        = {Introducing talkie: {A} 13{B} vintage language model from 1930},
  author       = {Levine, Nick and Duvenaud, David and Radford, Alec},
  year         = {2026},
  month        = apr,
  howpublished = {Blog post},
  url          = {https://talkie-lm.com/introducing-talkie}
}

@article{luo2026pretraininghistorical,
      title={Pretraining Language Models on Historical Text}, 
      author={Xiaoxi Luo and Zachary Shinnick and Niclas Griesshaber and Yixuan Wang and Junchi Yu and Freda Shi and Philip Torr and Yao Lu},
      year={2026},
      journal = {arXiv preprint arXiv:2606.02991},
      eprint={2606.02991},
      archivePrefix={arXiv},
      primaryClass={cs.CL},
      url={https://arxiv.org/abs/2606.02991}, 
}

@article{AI2reasoning2018,
  author       = {Peter Clark and
                  Isaac Cowhey and
                  Oren Etzioni and
                  Tushar Khot and
                  Ashish Sabharwal and
                  Carissa Schoenick and
                  Oyvind Tafjord},
  title        = {Think you have Solved Question Answering? {T}ry {ARC}, the {AI2} Reasoning
                  Challenge},
journal={arXiv preprint arXiv:1803.05457},
  volume       = {abs/1803.05457},
  year         = {2018},
  eprinttype   = {arXiv},
  eprint       = {1803.05457},
  bibsource    = {dblp computer science bibliography, https://dblp.org}
}

@article{MMLU2020,
  author       = {Dan Hendrycks and
                  Collin Burns and
                  Steven Basart and
                  Andy Zou and
                  Mantas Mazeika and
                  Dawn Song and
                  Jacob Steinhardt},
  title        = {Measuring Massive Multitask Language Understanding},
  journal={arXiv preprint arXiv:2009.03300},
  volume       = {abs/2009.03300},
  year         = {2020},
  url          = {https://arxiv.org/abs/2009.03300},
  eprinttype   = {arXiv},
  eprint       = {2009.03300},
  bibsource    = {dblp computer science bibliography, https://dblp.org}
}

@article{zhou2023instructionfollowing,
  title={Instruction-Following Evaluation for Large Language Models},
  author={Zhou, Jeffrey and Lu, Tianjian and Mishra, Swaroop and Brahma, Siddhartha and Basu, Sujoy and Luan, Yi and Zhou, Denny and Hou, Le},
  year={2023},
  journal = {arXiv preprint arXiv:2311.07911},
  eprint={2311.07911},
  archivePrefix={arXiv},
  primaryClass={cs.CL}
}

@article{lin2026jptlbenchanchoredpairwisellm,
      title={{JP-TL}-Bench: Anchored Pairwise {LLM} Evaluation for Bidirectional {J}apanese-{E}nglish Translation}, 
      author={Leonard Lin and Adam Lensenmayer},
      year={2026},
      eprint={2601.00223},
      archivePrefix={arXiv},
      journal={arXiv preprint arXiv:2601.00223},
      primaryClass={cs.CL},
      url={https://arxiv.org/abs/2601.00223}, 
}

@inproceedings{li2025arenahard,
  title     = {From Crowdsourced Data to High-Quality Benchmarks:
               {A}rena-{H}ard and {B}ench{B}uilder Pipeline},
  author    = {Li, Tianle and Chiang, Wei-Lin and Frick, Evan and
               Dunlap, Lisa and Wu, Tianhao and Zhu, Banghua and
               Gonzalez, Joseph E. and Stoica, Ion},
  booktitle = {Proceedings of the 42nd International Conference on Machine Learning},
  pages     = {34209--34231},
  year      = {2025},
  volume    = {267},
  series    = {Proceedings of Machine Learning Research},
  publisher = {PMLR},
  url       = {https://proceedings.mlr.press/v267/li25h.html}
}

@inproceedings{mozafari2026qdaps,
  title     = {Question Difficulty Estimation for Large Language Models
               via Answer Plausibility Scoring},
  author    = {Mozafari, Jamshid and Piryani, Bhawna and Jatowt, Adam},
  booktitle = {Proceedings of the 64th Annual Meeting of the
               Association for Computational Linguistics
               (Volume 1: Long Papers)},
  pages     = {11124--11151},
  year      = {2026},
  address   = {San Diego, California, United States},
  publisher = {Association for Computational Linguistics},
  doi       = {10.18653/v1/2026.acl-long.510},
  url       = {https://aclanthology.org/2026.acl-long.510/}
}

@inproceedings{don-yehiya2026mediocrity,
  title     = {Mediocrity is the key for {LLM} as a Judge Anchor Selection},
  author    = {Don-Yehiya, Shachar and Yehudai, Asaf and
               Choshen, Leshem and Abend, Omri},
  booktitle = {Proceedings of the 64th Annual Meeting of the
               Association for Computational Linguistics
               (Volume 1: Long Papers)},
  pages     = {15491--15513},
  year      = {2026},
  address   = {San Diego, California, United States},
  publisher = {Association for Computational Linguistics},
  doi       = {10.18653/v1/2026.acl-long.706},
  url       = {https://aclanthology.org/2026.acl-long.706/}
}

@inproceedings{west2024generative,
  title     = {The Generative {AI} Paradox: ``What It Can Create, It May Not Understand''},
  author    = {Peter West and Ximing Lu and Nouha Dziri and Faeze Brahman
               and Linjie Li and Jena Hwang and Liwei Jiang and Jillian Fisher
               and Abhilasha Ravichander and Khyathi Chandu and Benjamin Newman
               and Pang Wei Koh and Allyson Ettinger and Yejin Choi},
  booktitle = {The Twelfth International Conference on Learning Representations},
  year      = {2024},
  url       = {https://openreview.net/forum?id=CF8H8MS5P8}
}

@misc{breen2026vintage,
  author       = {Benjamin Breen},
  title        = {Are ``Vintage {LLM}s'' the Start of a New Humanistic Field? {T}houghts on Historical Language Models and {Talkie-1930}},
  year         = {2026},
  month        = apr,
  day          = {29},
  howpublished = {\textit{Res Obscura}},
  url          = {https://resobscura.substack.com/p/are-vintage-llms-the-start-of-a-new}
}

@article{bradley1952rank,
  author  = {Bradley, Ralph Allan and Terry, Milton E.},
  title   = {Rank Analysis of Incomplete Block Designs: I. {T}he Method of Paired Comparisons},
  journal = {Biometrika},
  volume  = {39},
  number  = {3/4},
  pages   = {324--345},
  year    = {1952},
  doi     = {10.2307/2334029}
}

@article{cattelan2012models,
  author  = {Cattelan, Manuela},
  title   = {Models for Paired Comparison Data: A Review with Emphasis on Dependent Data},
  journal = {Statistical Science},
  volume  = {27},
  number  = {3},
  pages   = {412--433},
  year    = {2012},
  doi     = {10.1214/12-STS396}
}

@article{hunter2004mm,
  author  = {Hunter, David R.},
  title   = {{MM} Algorithms for Generalized {Bradley--Terry} Models},
  journal = {The Annals of Statistics},
  volume  = {32},
  number  = {1},
  pages   = {384--406},
  year    = {2004},
  doi     = {10.1214/aos/1079120141}
}

@article{gelman2008weakly,
  author  = {Gelman, Andrew and Jakulin, Aleks and Pittau, Maria Grazia and Su, Yu-Sung},
  title   = {A Weakly Informative Default Prior Distribution for Logistic and Other Regression Models},
  journal = {The Annals of Applied Statistics},
  volume  = {2},
  number  = {4},
  pages   = {1360--1383},
  year    = {2008},
  doi     = {10.1214/08-AOAS191}
}

@article{pollitt2012comparative,
  author  = {Pollitt, Alastair},
  title   = {Comparative Judgement for Assessment},
  journal = {International Journal of Technology and Design Education},
  volume  = {22},
  number  = {2},
  pages   = {157--170},
  year    = {2012},
  doi     = {10.1007/s10798-011-9189-x}
}

@article{pollitt2012method,
  author  = {Pollitt, Alastair},
  title   = {The Method of Adaptive Comparative Judgement},
  journal = {Assessment in Education: Principles, Policy \& Practice},
  volume  = {19},
  number  = {3},
  pages   = {281--300},
  year    = {2012},
  doi     = {10.1080/0969594X.2012.665354}
}

@article{hamilton2026parameter,
  author  = {Hamilton, Ian and Tawn, Nick},
  title   = {Parameter Estimation in Comparative Judgment Under Random and Adaptive Scheduling Schemes},
  journal = {Journal of Educational Measurement},
  volume  = {63},
  number  = {1},
  pages   = {e70022},
  year    = {2026},
  doi     = {10.1111/jedm.70022}
}

@inproceedings{zheng2023judging,
  author    = {Zheng, Lianmin and Chiang, Wei-Lin and Sheng, Ying and Zhuang, Siyuan
               and Wu, Zhanghao and Zhuang, Yonghao and Lin, Zi and Li, Zhuohan
               and Li, Dacheng and Xing, Eric P. and Zhang, Hao
               and Gonzalez, Joseph E. and Stoica, Ion},
  title     = {Judging {LLM}-as-a-Judge with {MT-Bench} and {Chatbot Arena}},
  booktitle = {Advances in Neural Information Processing Systems},
  volume = {36},
  pages     = {46595--46623},
  year      = {2023}
}

@article{ford1957solution,
  author  = {Ford, Jr., L. R.},
  title   = {Solution of a Ranking Problem from Binary Comparisons},
  journal = {The American Mathematical Monthly},
  volume  = {64},
  number  = {8P2},
  pages   = {28--33},
  year    = {1957},
  doi     = {10.1080/00029890.1957.11989117}
}

@incollection{platt1999probabilistic,
  author    = {Platt, John C.},
  title     = {Probabilistic Outputs for Support Vector Machines and
               Comparisons to Regularized Likelihood Methods},
  booktitle = {Advances in Large-Margin Classifiers},
  editor    = {Smola, Alexander J. and Bartlett, Peter L. and
               Sch\"{o}lkopf, Bernhard and Schuurmans, Dale},
  publisher = {MIT Press},
  address   = {Cambridge, MA},
  pages     = {61--74},
  year      = {1999}
}

@book{cizek2007standard,
  author    = {Cizek, Gregory J. and Bunch, Michael B.},
  title     = {Standard Setting: A Guide to Establishing and Evaluating
               Performance Standards on Tests},
  publisher = {Sage},
  address   = {Thousand Oaks, CA},
  year      = {2007}
}

@article{davies2012coha,
  author  = {Davies, Mark},
  title   = {Expanding Horizons in Historical Linguistics with the
             400-Million Word {Corpus of Historical American English}},
  journal = {Corpora},  year = {2012},  volume = {7},  number = {2},
  pages   = {121--157},  doi = {10.3366/cor.2012.0024},
}

@misc{eccotcp,
  author       = {{Text Creation Partnership}},
  title        = {Eighteenth Century Collections Online, Text Creation
                  Partnership ({ECCO-TCP})},
  year         = {2011},  organization = {University of Michigan Library},
  howpublished = {\url{https://textcreationpartnership.org/tcp-texts/ecco-tcp-eighteenth-century-collections-online/}},
  note         = {Accessed 2026-09-09},
}

@misc{oapen,
  author       = {{OAPEN Foundation}},  title = {{OAPEN L}ibrary},
  year         = {2011},  address = {The Hague, Netherlands},
  howpublished = {\url{https://www.oapen.org/}},
  note         = {Open-access repository of peer-reviewed scholarly books;
                  accessed 2026-09-09},
}

@article{matheson1976,
  author  = {Matheson, James E. and Winkler, Robert L.},
  title   = {Scoring Rules for Continuous Probability Distributions},
  journal = {Management Science},
  year    = {1976},
  volume  = {22},
  number  = {10},
  pages   = {1087--1096},
  doi     = {10.1287/mnsc.22.10.1087},
}

@article{gneiting2007,
  author  = {Gneiting, Tilmann and Raftery, Adrian E.},
  title   = {Strictly Proper Scoring Rules, Prediction, and Estimation},
  journal = {Journal of the American Statistical Association},
  year    = {2007},
  volume  = {102},
  number  = {477},
  pages   = {359--378},
  doi     = {10.1198/016214506000001437},
}

@article{vallender1974calculation,
  title   = {Calculation of the {Wasserstein} Distance Between
             Probability Distributions on the Line},
  author  = {Vallender, S. S.},
  journal = {Theory of Probability \& Its Applications},
  volume  = {18},
  number  = {4},
  pages   = {784--786},
  year    = {1974},
  doi     = {10.1137/1118101}
}

@inproceedings{boerner2023access,
  author    = {Boerner, Timothy J. and Deems, Stephen and Furlani, Thomas R.
               and Knuth, Shelley L. and Towns, John},
  title     = {{ACCESS}: Advancing Innovation: {NSF}'s Advanced
               Cyberinfrastructure Coordination Ecosystem: Services \& Support},
  booktitle = {Practice and Experience in Advanced Research Computing},
  series    = {PEARC '23},
  year      = {2023},
  month     = jul,
  pages     = {173--176},
  publisher = {ACM},
  address   = {New York, NY, USA},
  doi       = {10.1145/3569951.3597559},
  url       = {https://doi.org/10.1145/3569951.3597559}
}

\appendix
\section{Scoring freely-generated answers}
\label{sec:appendix-scoring}

In automated scoring of free generation, every question gets a substantive score (through one of two pathways), as well as several stylistic scores. (See Figure \ref{fig:scoring-pipeline}.) Here we cover aspects of the scoring process too involved for the main text.

\subsection{Stylistic scoring}
\label{sec:appendix-style-score}

We use the word ``style'' to describe a wide range of linguistic differences between texts, involving diction, syntax, or tone. We do not intend to imply that these differences are only cosmetic. On the contrary, they are often consequential.

For instance, what is the cause of blood flow in the human body? An answer from an 1888 medical textbook: ``The real cause of the flow is the ventricular stroke.'' Compare the answer provided by a contemporary language model: ``The primary driver of blood flow is the heart's contraction and relaxation.'' The second sentence may seem to convey the same meaning, without obvious anachronism. All of these words existed in 1888. But ``real cause'' was much more common in the nineteenth century, and ``primary driver'' is rare before 1960. The difference between the two is subtle, but not cosmetic: one of the phrases implies a monocausal explanation and the other carefully avoids it. 

Tacit implications of this kind are important in intellectual history, but they can be hard to perceive and interpret. We cannot expect human readers, or LLM judges, to catch and explain all of them. Instead, we have trained discriminative judges to measure degrees of linguistic difference. When a model's imitation of the past veers off target, we should be able to measure the discrepancy, even if we can't say in every case why it matters. Our measurements can be divided into two categories: date prediction and authenticity detection.

\subsubsection{Date predictor}

The training corpus here is drawn from a roster of 4,325 human-written texts spread across the timeline from 1700 to 2020, with benchmark texts excluded. We are primarily concerned to cover the core century from 1831–1930. So we aim for 150 volumes per decade in that period. But the shoulders of the timeline also matter, since language models might accidentally produce 21st-century-sounding text---or overdo the archaism and hit an 18th-century target. We aim for 100 volumes a decade in shoulder periods. Sources include the Institutional Data Initiative, ECCO, COHA, OAPEN, and the Chicago Novel Corpus \citep{idi_institutional_books_1_2024, eccotcp, davies2012coha, oapen}. We strive to keep genres roughly balanced across time.

From this roster we sample 66,000 passages, exactly 2,000 per decade from the 1700s to the 2020s, with a length distribution matched to the length distribution of real benchmark answers. (Fragments of less than a single sentence are excluded from date scoring.) Train, validation, and test splits are grouped by volume and author, so no author appears on both sides of a split.

We tried several different instruments. The one we settled on is DeBERTa-v3-large with a 36-way softmax over ten-year bins spanning 1680–2040, fine-tuned at a maximum length of 256 tokens. Ordinality enters through the training target---a passage published in 1883 is trained not against a one-hot label but against a Gaussian bump centered on 1883 with $\sigma$ = 15 years---so that missing by one bin is penalized far less than missing by ten. Bins extend two decades beyond the corpus at each end so that the Gaussian target is not truncated at the extremes---since the target is renormalized, truncation would bias edge labels inward.

Since we are predicting a distribution, and evaluating it against a single date, we measure error by calculating a Continuous Ranked Probability Score \citep{matheson1976, gneiting2007}, which rewards the model both for accuracy and for precision. For held-out texts in the core 1831-1930 century that concerns us, this measure of error is 16.7 years for a single answer.

We don't use date predictions as a direct verdict. Knowing that a passage supposed to be written in 1880 was scored at 1907 doesn't tell us much until we know how actual passages from 1880 get scored in this model. So we use date predictions conformally: each candidate answer is ranked against a matched sample of authentic period prose that has been scored in the same way. To achieve this we construct a reference corpus  by assembling roughly 43,000 passages of genuine 1831–1930 prose (with margins out to 1821–1940). Passages are divided into six length bins, because longer passages can be scored with more precision.

For each candidate answer we build a comparison window from the reference corpus: passages published within ±10 years of the date the question targets, and falling in the same length bin. We compare the signed residual of the candidate answer---predicted mean date minus target date---against the window's own residuals, to place the answer as a percentile.

If a model's answers were stylistically indistinguishable from matched authentic prose, the percentiles of its answers would be drawn from
Uniform(0, 1). So we summarize the date channel by the exact Wasserstein-1 distance of the model's percentile distribution from uniform \citep{vallender1974calculation}. However, even genuine period text produces a nonzero distance from uniform at finite sample size, so our final conformal move is to generate 2000 pseudo-models: draws from the reference corpus matched to the candidate model's distribution across dates and length bins. Their mean distance from uniform, $W_0$, is the baseline a model with
nothing wrong with it would post. The headline scores rescale the observed distance against that baseline onto 0–100, so that date fidelity is
\begin{equation}
f = 100 \cdot (1 - \max(0, W - W_0) / (0.5 - W_0))
\end{equation}
The upshot of this for interpretation is that we are never just comparing the mean average error of a candidate model to a null model of mean average error. Since errors vary, each passage is compared against its own date and length bin, and when we back up to the model level the distribution of errors also matters. A candidate model with no error on half its passages, and twice average error on the other half, would not be indistinguishable from null.

\subsubsection{Human authenticity detector}

Language models can also fail to sound historically persuasive, not because they sound more modern or more archaic than the intended target, but simply because they sound like language models.

To measure this failure mode, we train a second DeBERTa model to distinguish authentic human text (in the century of interest) from LLM imitations. Our authentic corpus is 29,000 passages drawn from IDI texts in the period (excluding volumes used for the benchmark). To create imitations, we developed a stable of models that included base, instruction, and reasoning models---even Talkie-1930. We allowed these models to base their responses on authentic templates in a variety of ways: paraphrase, infill, continuing a passage, even constrained-generation and few-shot prompting. We stripped punctuation that might give away provenance, and ensured both negative and positive examples matched the length distribution of benchmark answers $\geq 5$ words long.

On a held-out test split, our DeBERTa model achieved accuracy 0.881, F1 0.875, AUROC 0.941. On held-out models (including held-out model families) it achieved accuracy 0.901, F1 0.901, AUROC 0.964. We undertook the same sort of conformal calibration here that was used for the date predictor, using the same reference corpus. But instead of using a signed residual to rank the candidate answer within its ±10 year window, we used the predicted probability of authenticity.

Candidly, however, we advise against giving this measure much weight, since it seems relatively easy to improve this score through fine-tuning.

\subsection{Substantive scoring}
\label{sec:appendix-substance-score}
In general the goal of substantive scoring is to assess whether a model's answer follows instructions, is accurate, and fits the specified social context. ``Fitting the context'' can have a rhetorical dimension, so stylistic and substantive measurement should not be envisioned as an exclusive opposition, but as two intersecting Venn circles. Style is often part of substance. If one wanted to define an exclusive opposition, one could oppose the discriminative, fine-tuned judges of section \ref{sec:appendix-style-score} to the general-purpose generative judges of section \ref{sec:appendix-substance-score}. 

\subsubsection{Pass-fail scoring}

The substantive dimension of some questions is straightforward enough that they can be scored simply by comparing a candidate answer to ground truth. We ask a judge to compare the two without labels, and choose the answer that is better on ``factual accuracy (in the specified historical context) and relevance to the question (correctly following instructions).'' If ``both answers are accurate and relevant,'' the judge can indicate a tie.

Since it is not really meaningful for an answer to exceed ground truth, a tie is the significant threshold here. But we construct an unlabeled comparison, instead of directly asking whether an answer is as good as (labeled) ground truth, because a judge's \emph{failure to discriminate} between unlabeled options can be tested more reliably than its characterization of labeled ones.

We can use exactly the same prompt and question framing, for instance, but ask the judge to discriminate between ground truth and known (human-labeled) distractors. This gives us an estimate of $\alpha$, the frequency with which the judge fails to reject answers that should be rejected. We estimate this per-question, testing each ground truth answer against all distractors that would be wrong on the listed criteria. (Purely stylistic failures are not significantly wrong for this task, so those distractors are not used.) We test each comparison using both answer orderings, and count the number of failures to reject (a tie or preference for the distractor both count as failure).

Questions with $\alpha > 0.25$ do not get pass-fail scoring; they are moved to the more sensitive and time-consuming partial-credit path. Pooling trials among the 520 questions that remain, and measuring the proportion of failures, $\alpha$ for Claude Sonnet 4.6 is $0.027$.

Because we have multiple ground-truth options for 72 questions in this scoring pathway, we can also estimate $\beta$, the frequency with which a judge wrongly rejects a valid answer. We test this by offering two ground truth answers (in both possible sequences) and measuring the frequency with which the judge prefers one or the other, rather than reporting a tie, which is by construction the correct choice.

In this question framing, either preference is a material error, but in real evaluation, preferring a (valid) answer over ground truth has no effect---it is equivalent to choosing a tie. So we interpret the raw error rate in testing as $2\beta$, and halve it to get $\beta = .042$. Both error rates are thus below 5\%.

In actual scoring of candidate models on this pathway, we do not use both possible sequences; we choose one randomly and allow the random bias to spread across 520 questions. In cases where a question has two possible ground truth answers, we test both, and also vary the ordering once to break the tie, if one emerges---as a result there are (very rare) cases where pass/fail scoring can produce a score of 0.33 or 0.66.

\subsubsection{Partial-credit scoring}

The hardest aspect of judging this benchmark is judging whether an answer fits a given socio-historical context. The stylistic dimension of that question can, to some extent, be outsourced to discriminative models; see section \ref{sec:appendix-style-score}. But there are also kinds of failure a style judge will miss, like ``this was said at the time, but it would have been unusual to hear it from an Anglican clergyman.''

We tried a number of solutions here, including manual human scoring, but finally concluded that 346 questions---including all ``generation'' questions---entail an extensive gray area that is not well represented as a single pass/fail choice (even with expert human judges). Instead, we create a graduated scale, following the logic of Comparative Judgement in educational assessment, where pairwise comparisons of student work are fit to yield a per-task quality scale \cite{pollitt2012comparative, pollitt2012method}.

As anchors for that scale, we use the answer options created for likelihood scoring (both ground truth answers and distractors). We stage pairwise forced-choice comparisons between each pair of answers, asking specifically which best fits the historical context. We infer latent strengths for all options as outlined in \citet{bradley1952rank, cattelan2012models}. Since we are looking at relatively small graphs (5-8 vertices and 10-28 edges for each question), it frequently happens that all answers in one subgraph defeat all answers in the other. Under this condition the maximum likelihood estimate is not finite \citep{ford1957solution, hunter2004mm}. \citet{hamilton2026parameter} address the same problem in Comparative Judgement with likelihood penalties; we use the Bayesian equivalent, a weakly informative prior \citep{gelman2008weakly}. To balance order biases, we repeat every comparison twice, with both orderings, as recommended in \citet{zheng2023judging}. Each trial is treated as a separate comparison; conflicting results are not collapsed into a tie. We use Claude Sonnet 5 as a judge for the partial-credit path; we could also have used it for pass/fail scoring, but there is no reason to enforce uniformity.

\begin{figure*}[ht]
    \centering
    \includegraphics[width=\textwidth]{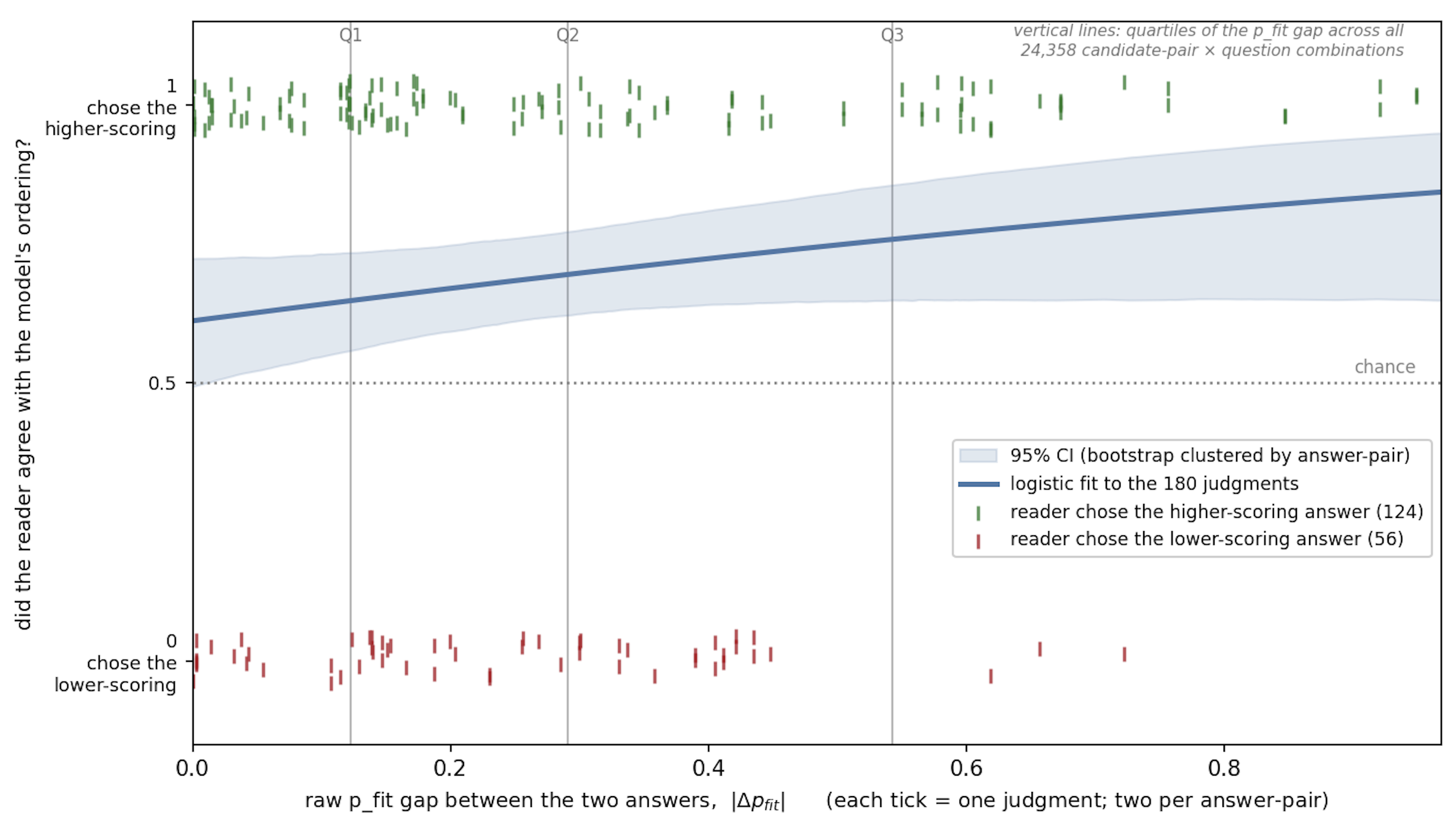}
    \caption{Observed directional agreement between six human readers and our Bradley-Terry scoring model. Gaps between the two answers are displayed here on the [0,1] scale of $p_{fit}$, although the logistic model is actually fit to $\theta$, an unbounded quantity.}
    \label{figure:readermodel}
\end{figure*}

The latent strengths ($\theta$s) thereby inferred are treated as frozen anchors for each question. Candidate answers are then scored through pairwise comparisons to all the anchors, repeating each comparison twice to vary the answer order. We measure

\begin{equation}
  \Delta_\theta = \theta_{c} - \bar{\theta}_{\mathrm{GT}}
\end{equation}

\noindent and take this as a measure of the answer's distance from ground truth.

The last stage of this process is to scale $\Delta_\theta$ and make it comparable to other parts of the benchmark. We do that through LOO scoring of held-out ground-truth answers and distractors, which gives us independent $\Delta_\theta$ estimates paired with answers that are known to fit, or not fit, the historical context. (To ensure a large gap does not emerge between distractors and ground truth, we label some strong distractors as partly acceptable, but the dominant scaling evidence---23 times more influential--is provided by variation among ground truth answers.) We fit a logistic model to infer the probability that an answer with a given $\Delta_\theta$ fits the context ($p_{fit}$) \citep{platt1999probabilistic}. The intercept is pinned so that the average ground truth answer ($\Delta_\theta = 0$) receives a score of 0.9. This is what \citet{cizek2007standard} call a ``standard-setting'' choice; without this intervention the intercept would end up considerably lower (0.33-0.65). We intervene, and move ground truth up to 0.9, explicitly in order to create a scale more comparable to other parts of the benchmark. We cannot move the intercept all the way to 1.0 (a finite input cannot produce 1.0), so some choice is necessary. And while we don't believe ``better than ground truth'' is a meaningful model-level goal, it might harm discriminative power to completely flatten answer-level variation above $\bar{\theta}_{\mathrm{GT}}$. 

\section{Human validation}
\label{sec:appendix-human}

In the binary scoring pathway, automated judgments are validated by asking the judge to choose between two known ground truths, or between a ground truth and a distractor that human beings have labeled as wrong. An anchor in human judgment is implied by question design, and as noted above, judge error rates on both tasks are < 5\%.

The task of validating partial-credit scoring is more challenging. Here we are interested in shades of gray. If human experts could confidently point to answers that are 30\%, 50\%, or 70\% likely in a specified historical context, we could use those graduated labels to calibrate a scale. But human experts do not actually have this level of confidence when an answer is considered in isolation. Instead, as noted above, we use human-crafted judgment rubrics to infer latent strengths $\theta$ through pairwise comparisons. (Then we calibrate those to produce $p_{fit}$ in a separate step, using variation among ground truth answers as one key source of evidence for our calibration scale.)

But do automated judges, guided by human criteria, actually produce a ranking that corresponds to human preference? This is a question that requires separate validation. So we selected 90 questions, and in each case chose a pair of answers ($a$ and $b$) from two models that had taken our benchmark, distributing pairs across two axes: the median strength of $\theta$ across both answers, and the gap between $\theta_a$ and $\theta_b$. We then asked six human judges to make a forced-choice ranking of $a$ and $b$ in these same 90 pairs, organizing judgments so each judge saw the full range of pair difficulty, and each pair of answers was evaluated by two humans.

We were interested in the degree of agreement between the model and human judges---but also, and more centrally, in the way that degree of agreement varied across different sizes of gap between $a$ and $b$. After all, if $\theta$ and $p_{fit}$ are identical for the two answers, we would expect no reason for preference and no agreement beyond chance between our model and a human reader. So the absolute frequency of agreement is less important here than its distribution, and the level of discrimination implied. For that reason, we oversampled difficult cases with a/b gaps below the median, in order to improve resolution where it counts.

Each pair of answers was shown to two judges so we could also assess the difficulty of this task for human readers. They agreed in two-thirds ($66.7\%$) of cases. Overall agreement between human readers and our Bradley-Terry model was just slightly higher ($68.9\%$). Since we oversampled difficult cases, these figures should be interpreted cautiously. But it is notable that model-human agreement is statistically indistinguishable from human-human. A logistic model was trained to predict human preference from both the gap between answers ($\Delta_\theta)$ and the midpoint of the two. Using draws from that model and resampling of benchmark questions, we inferred $95\%$ confidence intervals for human preference for all 78 pairings of 13 models we had scored. We find that every $0.1$ of $p_{fit}$ difference between two models implies a 5\% human preference for A over a coin flip---or put another way, a 10\% preference gap between the two models. When the difference of $p_{fit}$ reaches $0.11$ our $95\%$ interval on human preference no longer includes $0.5$. Differences of that size are statistically significant.

\begin{table}[hb!]
\centering
\begin{tabular}{lrr}
\toprule
Distractor class & Likelihood & Free generation \\
\midrule
Partially valid & 0.64 & 0.69 \\
Negation & 0.17 & 0.44 \\
LLM-generated & 0.55 & 0.35 \\
Same character & 0.23 & 0.33 \\
Same book & 0.28 & 0.24 \\
Hand-written & 0.39 & 0.24 \\
Real source & 0.37 & 0.20 \\
\bottomrule
\end{tabular}
\caption{Rate at which each class of distractor defeats the model. Rates are conditional on the class being present in the question, and weighted equally per question. Distractors the evaluated model itself generated are excluded.}
\label{tab:distractor-difficulty-0.8}
\end{table}

\section{Error analysis}

When models make errors, what is the source of error? We can assess this by looking at cases where a model ``loses'' to a distractor---either (in likelihood scoring) because it perceives the distractor as more probable than ground truth, or (in partial-credit scoring of freely generated answers) because a judge prefers the distractor to its answer. These are different kinds of failure, but both cast light on distractor strength.

To permit apples-to-apples comparison of likelihood and free generation, we focus here on 346 partial-credit questions, which are the only ones that receive pairwise comparison to distractors in free generation. Results are presented in Table \ref{tab:distractor-difficulty-0.8}.

The most significant result here is that the ``partially valid'' answers---marked in advance by a human as the strongest distractors, and given a probability $0 < p < 1$---live up to their label. In free-generation scoring, this is not completely independent confirmation, because the answer rubric may imply that the kinds of errors typified by these distractors are less problematic than other errors. But there is no rubric for likelihood scoring, and these distractors are still able to defeat models about two-thirds of the time. This provides reassuring independent evidence that our human question designers can assess distractor difficulty.

Most of the ``partially valid'' distractors are LLM-generated, and other LLM-generated answers are also among the stronger distractors. There are also some categories that are challenging in likelihood scoring (hand-written by a human or drawn from another real historical source), but not particularly strong against freely generated candidates. Conversely, negations of ground truth are easy to reject in likelihood scoring, but sometimes present a challenge in a pairwise comparison.

\begin{table*}[!htp]
\centering
\footnotesize
\setlength{\tabcolsep}{4pt}
\begin{tabular}{@{}lp{0.11\textwidth}rrp{0.35\textwidth}p{0.29\textwidth}@{}}
\toprule
Item & Answer source & Prob. & $\theta$ & Answer & Reject reason \\
\midrule
\texttt{gt1} & \raggedright\texttt{ground truth} & 1.00 & +4.33 & The rifled gun, from its great accuracy and penetrating power, and more especially for its power of throwing a loaded shell with an accuracy and range beyond what have heretofore been considered to belong to solid-shot guns, is a very important gun to sea-coast batteries. Its most important use, however, will be as an incendiary and explosive agent against wooden vessels at long\,\ldots{} & \textit{--- accepted} \\
\addlinespace[3pt]
\texttt{gt0} & \raggedright\texttt{ground truth} & 1.00 & +1.40 & In reference to rifle guns, it is doubtless true that they are destined to play an important part hereafter; that they possess qualities of range and accuracy which must make them, where these\,\ldots{} & \textit{--- accepted} \\
\addlinespace[3pt]
\texttt{d3} & \raggedright\texttt{claude-\allowbreak{}opus-\allowbreak{}5} & 0.50 & +1.34 & It is not to be concluded from what has been said above of the range and precision of the rifled gun that the spherical projectile has served its turn and may be laid aside; for the problem which the channels of our harbors impose upon the engineer is not that of striking a distant object with nicety, but of disabling, at ranges of a few hundred yards, a vessel cased in four inches and a half of wrought iron laid upon a solid backing of\,\ldots{} & is actually quite close to the perspective of an American engineer in 1861, and gives itself away only, perhaps, in its condensation: it reads suspiciously like someone has understood the whole argument and reverse-engineered an ideal paragraph expressing Barnard's dialectical perspective on rifled and smooth-bore artillery \\
\addlinespace[3pt]
\texttt{d1} & \raggedright\texttt{gemini-\allowbreak{}3.6-\allowbreak{}flash} & 0.00 & -1.61 & While the superior accuracy, range, and armor-penetrating power of elongated projectiles make rifled artillery destined to eventually supersede smooth-bore ordnance, heavy smooth-bores such as the 15-inch Rodman gun still serve a vital temporary role in coast defense. Their immense caliber\,\ldots{} & knows too much about the future, and in particular is too confident that rifled guns will eventually triumph; that, combined with the relatively modern syntax, suggests that this passage may have been composed by a 21c LLM not an 1861 military engineer \\
\addlinespace[3pt]
\texttt{d2} & \raggedright\texttt{1873 text} & 0.00 & -2.33 & In batteries liable to attack by iron-plated ships, the heaviest rifled guns, ML. 7 to 12-in., should be mounted; but against wooden vessels the BL. R. 7-in. and 64-pr., the ML. R. 80 and 64-prs.,\,\ldots{} & uses British terminology, referring to guns by the weight of the projectiles they throw, and also seems to be written later than 1861; it can already take for granted the existence of very large rifled artillery pieces whose viability was still doubtful in 1861 \\
\addlinespace[3pt]
\texttt{d0} & \raggedright\texttt{1864 text} & 0.00 & -3.12 & The experiments were concluded by firing a few rounds from the 110-pr. at the field splinter proof which had hitherto sheltered the range party during the practice: it was struck three times in six\,\ldots{} & is clearly written from the context of practical siege warfare and field engineering, where the problem is breaching earthen fortifications, rather than from Barnard's seacoast-defense context, where the central problem is stopping ships and armored vessels \\
\bottomrule
\end{tabular}
\caption{Answer options for question 888 (Based on \emph{Notes on Sea-coast Defence,} J. G. Barnard, 1861). $\theta$ is the posterior mean latent strength from the frozen Bradley--Terry anchor fit, so rows are ordered by the judge's own ranking; probability is the benchmark's answer probability, where $0.5$ marks a distractor human editors judged partly valid.}
\label{tab:answers-q888-0.8}
\end{table*}

\subsection{Example of Bradley-Terry scoring}
\label{sec:appendix-bt-example}

A worked example of scoring for a partial-credit question may be useful. In Table \ref{tab:answers-q888-0.8}, we see the answer options for question \#888.

\textbf{Metadata frame:} The following question is based on a technical discussion by a U.S. Army engineer writing about the armament of seacoast fortifications in 1861. The ideal answer will resemble a passage that could have appeared in that context.

\textbf{Question:} One question explicitly posed in this document is: ``Of what use are the new larger smooth-bore guns—for example, those of 15 inches recently produced? Are not rifled guns destined to supersede all smooth-bore guns?'' What answer would you expect? Write a paragraph of three sentences that might appear in the specified context.

As you will see in the table, there are two ground truth options, a strong distractor (ranked just below ground truth), and three weaker distractors, one written by LLM and two taken from other texts in the period.

For an example of scoring, here's an answer provided by GPT-4o: ``The new 15-inch smooth-bore guns offer significant advantages on the battlefield, particularly in their ability to launch substantial explosive shells capable of considerable destructive effect against fortifications and vessels. While rifled guns, with their increased range and accuracy, present a promising future for artillery, the immense power and area effect of the larger smooth-bores cannot be dismissed, especially for coastal defense where broader dispersal of shot proves advantageous. Thus, each type serves distinct roles, the smooth-bore for area bombardment and the rifled for precision, suggesting that \emph{until rifled technology matures further,} both will coexist in our armament.'' (italics added)

This answer defeated the two weakest distractors (in both trials), but lost (in both trials) to d3 and d1. Reasons for its losses are not recorded, but the concluding proviso (``until rifled technology matures further'') would likely fall afoul of this stricture in the rubric that guides the judge (originally aimed at the flaws of d1): ``knows too much about the future, and in particular is too confident that rifled guns will eventually triumph.'' The GPT-4o answer received a $\theta$ of -1.65 and a $p_{fit}$ of 0.159, for rather weak partial credit.

\begin{table*}[ht!]
\centering
\footnotesize
\setlength{\tabcolsep}{4pt}
\begin{tabular}{llcccc}
\toprule
\textbf{Model} & \textbf{Effort} & \textbf{Cloze} & \textbf{Generation} & \textbf{Knowledge} & \textbf{Date fidelity} \\
\midrule
Talkie 1930 13B base &  & 1.5 [0.8, 3.3] & 12.6 [11.1, 20.6] & 20.3 [15.5, 25.3] & 54.4 [50.0, 59.3] \\
Talkie 1930 13B it &  & 6.3 [4.8, 9.5] & 32.9 [29.6, 41.8] & 22.6 [17.5, 27.9] & 87.7 [82.4, 92.4] \\
\addlinespace[4pt]
Qwen2.5 7B &  & 7.3 [5.4, 10.7] & 13.2 [12.3, 22.4] & 25.5 [19.7, 31.0] & 4.8 [3.5, 6.2] \\
Qwen2.5 72B it &  & 19.1 [15.6, 24.0] & 20.2 [18.3, 29.9] & 50.2 [43.6, 56.6] & 13.8 [11.4, 16.5] \\
\addlinespace[4pt]
Gemini 3.7 flash & high & 85.8 [82.0, 88.9] & 60.9 [54.9, 66.4] & 98.2 [96.6, 99.6] & 76.7 [71.1, 82.8] \\
\addlinespace[4pt]
Kimi K3 & max & 85.7 [81.9, 88.8] & 66.6 [59.9, 71.2] & 94.4 [91.5, 97.1] & 94.8 [89.1, 98.7] \\
\addlinespace[4pt]
GPT-4o 2024-08-06 &  & 35.0 [30.8, 40.5] & 26.0 [23.5, 36.0] & 65.0 [59.0, 70.9] & 19.6 [16.1, 23.4] \\
GPT-4.1 &  & 65.6 [60.6, 70.1] & 44.0 [39.5, 52.9] & 68.9 [62.8, 74.5] & 65.8 [60.5, 71.7] \\
GPT-4.1 finetune &  & 38.9 [34.2, 43.6] & 53.8 [48.2, 60.1] & 62.7 [56.4, 68.7] & 98.9 [94.4, 99.5] \\
GPT-5.4 & medium & 89.9 [86.5, 92.3] & 68.0 [61.2, 72.3] & 93.1 [89.7, 96.1] & 94.2 [88.7, 98.7] \\
GPT-5.4 wrong contexts & medium & 83.7 [79.7, 87.0] & 54.3 [48.9, 60.7] & 85.9 [81.2, 90.2] & 91.0 [85.5, 95.9] \\
GPT-5.6 & high & 91.0 [87.7, 93.3] & 71.7 [64.7, 75.2] & 92.3 [88.5, 95.7] & 92.3 [86.7, 98.0] \\
GPT-6-Astra & high & 93.0 [90.0, 94.7] & 71.4 [64.7, 75.1] & 96.1 [93.5, 98.3] & 99.3 [94.5, 100.0] \\
\bottomrule
\end{tabular}
\caption{95\% confidence intervals for the point estimates in Table~\ref{tab:free-gen-results}, on the same rows and in the same order. All four measures run 0--100, higher is better. Intervals are percentile intervals from the bootstraps described above; because the point estimate is a plug-in estimate and the interval a percentile of bootstrap replicates, the two are not concentric on the partial-credit channel (the Generation column), and the date-fidelity interval is one-sided where it reaches 100.0.}
\label{tab:free-gen-intervals}
\end{table*}

\section{Confidence intervals}
\label{sec:appendix-confidence}

In the free-generation path, uncertainty of substantive scoring is estimated by a bootstrap over 2,000 replicates. The two scoring channels contribute variance differently. On the pass/fail channel the score is the arithmetic mean of the observed judge verdicts $v_q$, with no correction for judge error, so the only thing resampled is which questions were asked. The partial-credit channel resamples three nested layers per replicate. 
\begin{itemize}
\item First, judgment: each question's partial credit rests on $\Delta_\theta$, the Bradley–Terry latent-quality gap between the candidate's answer and the mean of the ground-truth answers, and the pipeline stores 1,000 posterior draws of $\Delta_\theta$ per question, so a replicate draws one index into that posterior per question. 
\item Second, instrument: the map from $\Delta_\theta$ to a probability, $p_{fit} = \sigma(a + b\Delta_\theta)$, was fit by logistic regression on labelled leave-one-out comparisons and bootstrapped by resampling whole questions (a cluster bootstrap over 100 question-clusters, 1,000 replicates); each scoring replicate draws one calibration pair. \item Third, item: as on the other channel, questions are resampled with replacement. The two channels draw from independent random streams, and the pooled scores---count-weighted and equal-weighted---are recomputed within each replicate from the two channel means, so their intervals reflect both channels jointly. 
\end{itemize}

Date fidelity and human authenticity also carry 95\% percentile intervals, calculated in a separate bootstrap of 1,000 replicates grouped by question and by reference volume.

Confidence intervals for the free-generation path appear in Table \ref{tab:free-gen-intervals}.

\section{Sample questions and distractors}
\label{sec:appendix-questions}
\subsection{Inference}

\textbf{metadata frame:} The following question is drawn from \emph{A Text-book on Applied Mechanics: Specially Arranged for the use of Science and Art, City and Guilds of London Institute and Other Engineering Students,} a textbook published in 1895 by Andrew Jamieson, a Scottish professor of electrical engineering.

\textbf{main question:} The saddle of a lathe weighs 5 cwts., and it is moved along the bed of the lathe by a rack and pinion arrangement. What force, applied at the end of a handle 10 inches in length, will be just capable of moving the saddle, supposing the pinion to have 12 teeth of 1 1/4-inch pitch, and the coefficient of friction between the saddle and lathe-bed to be 0.1, other friction being neglected?

\textbf{ground truth:} 13.36 lbs

\textbf{distractors:} 13.4 lbf, 50 cwts, 24 kg, 50 lbs

\subsection{Knowledge}

\textbf{metadata frame:} The following question asks for information from an American encyclopedia published in 1897; your answer should reflect the state of knowledge and style of exposition current at the time.

\textbf{main question:} What title did Philip V hold before becoming king of Spain?

\textbf{ground truth:} Duke Of Anjou

\textbf{distractors:} Prince of Castille, Governor of Granada, the Duke of Orange, insufficient information

\subsection{Abstention}

\textbf{metadata frame:} In answering the following question, envision the perspective of \emph{Radio up to the Minute,} a hobbyist textbook published in 1924 by John R. Nilson, an American radio engineer. Your answer should reflect the knowledge likely available in that volume.

\textbf{main question:} What was the primary advantage of the ferrite rod antenna?

\textbf{ground truth:} insufficient information

\textbf{distractors:} compact size, high sensitivity (Q factor), directivity, longer and larger than other antennae

\textbf{comment:} The advantages are real, but ferrite rod antennae did not exist until the 1940s. The prompt for knowledge, inference, and abstention questions counsels models to reply ``insufficient information'' if the question could not have been answered in the specified historical context.

\subsection{Structured cloze}

\textbf{metadata frame:} The following passage comes from \emph{Argentina,} a work of history and social description published in 1910 by W. A. Hirst, an American writer.

\textbf{main question:} France makes steady progress, and Italy shows a large increase, as is only to be expected, because the emigration from Italy has long been very large. It may be added that

French goods make their way by sheer merit, for France has in her own land ample scope for her scanty population. Some advantage may be obtained by her as the head of the Latin race, [missing clause revising an implied expectation]. It is very interesting to see how Argentina has passed from small to great things in matters of trade.

Write a clause appropriate for this book that could stand in the position marked by [missing clause revising an implied expectation]:

\textbf{ground truth:} but wherever there are women and luxury there will French trade flourish, and further, in machinery of many kinds France, if equalled by any other nation for excellence, is equalled by England alone

\textbf{distractors:} but wherever there are scarcity and simplicity, French trade will wither, and further, in machinery of many kinds, France, if surpassed by any other nation for excellence, is surpassed by England alone

but it is a peculiarity about Argentina that the people hurry to institute a great export trade long before they think of supplying themselves adequately with an article

but the third view is naturally that of Argentina, which is every year receiving an abundant stream of white colonists to develop the industries which cry aloud for labour

but her influence is tempered by the fierce competition from other nations within Latin America, even as its resources are comparatively scarce

\subsection{Character modeling}

\textbf{metadata frame:} This question asks you to provide dialogue for a character in a book. The book in this case is \emph{Kin-da-shon's wife; an Alaskan story,} a novel with ethnographic elements published in 1892 by Carrie M. W. Willard, an American writer born in 1853.

\textbf{main question:} The character in question here is Yealh-neddy, who is described this way in the book:
``Yealh-neddy is easily pleased, but hard to satisfy.'', ``Yealh-neddy was more than ever arrogant, and more than once the light from his evil eyes seemed to smite the girl.''

At one moment in the book, Yealh-neddy loudly declares he has cut the thongs from Sha-hehe and requests her healing as the boat departs. Write a short passage of dialogue (1-3 sentences) that Yealh-neddy might speak in this situation:

\textbf{ground truth:} I have cut the thongs from Sha-hehe. Cover her and heal her, old woman, against the time I come again. I shall want her then.

\textbf{distractors:} No more eagles than we've talons for, let me tell you.

Get in, Usha, and take the paddle. I can spring in when it is off the sands.

I have severed Sha-hehe's thongs, and now I demand your healing, lest the sea claim her. 

Leave me be, or suffer my wrath.

Ho! I have cut the thongs from Sha-hehe. Now let her be made whole before the boat has passed from sight.

\subsection{Constrained generation}

\textbf{metadata frame:} The following question is based on \emph{The Theosophical Forum,} a religious periodical published in New York in 1903. The best answer will be consistent with the doctrines of theosophy and the expository style of 1903.

\textbf{main question:} What can be affirmed about the fate of the soul after death?

\textbf{ground truth:} The human personality, whether called the soul or the spirit, does not go out of the body at death, but retreats within the body, back into the germ from which that body grew, and from which, provided the ``deeds done in the body'' were not such as to destroy its vitality, there is ground to infer that another body will grow.

\textbf{distractors:} The vital principle of Probation is that man's immortal destiny is determined by his character at the moment of death.

The book, however, is not given up to a theoretical discussion of the spirituality and the immortality of the soul or the freedom and responsibility of man; these matters are rather incidental, or should we say are everywhere assumed rather than proven, while the attention of the reader is directed to the practical issues of education.

The life and immortality the Gospel brings to light, is by an earthly resurrection from the dead. On the supposition that the soul is immortal, and exists in a disembodied state, the Gospel of Christ is made void.

Nothing is known about the soul, and little can be affirmed about its destiny after death.

According to Theosophical doctrine, the immortal soul enters a state of celestial rest and assimilation in Devachan after death, where it reflects upon earthly experiences before reincarnating to pursue further spiritual evolution and divine perfection.

\textbf{comment}: The benchmark corpus includes multiple religious points of view, and some of the distractors here become ground truth for the same question in a different context. The last distractor is consistent with Theosophy, but framed too self-consciously to be drawn from the specified context.

\section{Reasoning models' diagnostic abilities}
\label{sec:appendix-diagnostic}

When working as judges, models are not required to provide rationales for accepting or rejecting answers. We have only a quantitative estimate of their performance: see Appendices \ref{sec:appendix-scoring} and \ref{sec:appendix-human}.

But in informal experiments we have presented models with lists of answers where authentic passages are mixed with distractors, and asked them to explain their reasoning. There is always something slightly theatrical and Holmesian about a language model's explanation of its own deductive processes, but frontier models are very good at distinguishing LLM-written distractors from real human ones, and provide rationales for judgment that are in any event plausible. See Table \ref{tab:ladies-home-magazine-rationales}, where Claude Opus 5 analyzes ground truth answers and two distractors from question \#733, based on \emph{Arthur's Ladies' Home Magazine}, Philadelphia, 1871. 

\begin{table*}[!htp]
\centering
\footnotesize
\setlength{\tabcolsep}{5pt}
\begin{tabular}{@{}p{0.38\textwidth}cp{0.53\textwidth}@{}}
\toprule
Candidate answer & Label & Diagnosis by Claude Opus 5 \\
\midrule
The sleeves of the new walking-dress are made moderately close at the shoulder, widening toward the elbow, and finished at the wrist with a narrow frill of lace, which gives a light and graceful effect.
&
(a)
&
Language model. The tell isn't any single anachronism; it's the absence of nouns a real fashion writer couldn't have avoided. Three clauses about a sleeve and not one fabric, not one trade name, not one named sleeve shape. Period copy names things---coat sleeve, demi-long, bishop, pagoda---because the reader is being told what to ask for. This sentence instead describes the silhouette from scratch, which is what you do when you don't have the vocabulary and are working from a mental image. And the shape it describes is incoherent: narrow at the shoulder, widening toward the elbow, then somehow a wrist frill---that's a pagoda sleeve grafted onto a cuffed one, i.e., the average of two decades rather than either. The closer, ``which gives a light and graceful effect,'' is the giveaway you'd expect: an evaluative summary clause with no referent, appended because the model has learned that period prose sounds appreciative. Real columns do say the effect is graceful, but they say it about a named thing.
\\
\addlinespace[5pt]

His shirt, which looked as if it never could have been whole and white, had more than half the sleeves torn away, and fell open in front for want of a collar, to say nothing of a button and buttonhole.
&
(b)
&
I'm keeping, though it's the one I'd argue about. It's fiction, not fashion copy, so the genre heterogeneity is jarring---but ladies' magazines of the 1870s were mostly fiction by page count, and reprinting English serial matter was standard. It has the Dickensian comic apposition (``to say nothing of a button and buttonhole''), so if it's an impostor it's an 1850s--60s British novel rather than a model output. What holds me is ``looked as if it never could have been whole and white''---the counterfactual pushed back past the garment's own history is a strange enough construction that I don't think it's synthetic. Models write ``looked as though it had never been white.'' That extra modal is a human doing something odd on purpose.
\\
\addlinespace[5pt]

Dress of blue pique, made with a plain gored skirt, scalloped on the bottom, and trimmed with white linen braid and pearl buttons, and a plain square-necked waist without sleeves, worn over a guimpe of white nainsook, finished at the neck and wrists with Valenciennes lace.
&
(c)
&
Is the anchor. Pique, gored skirt, linen braid, guimpe over a sleeveless square-necked bodice, nainsook, Valenciennes---this is a misses' or child's summer dress plate, mid-1870s, and American. ``Waist'' for bodice is the decisive nationality marker; a British book would say bodice or corsage. The description is also usefully boring: it enumerates and stops, with no aesthetic verdict at the end.
\\
\addlinespace[5pt]

Pelisse of cotton-blossom-yellow gros de Naples, trimmed round the border, and on each side of the front, with vandykes in satin, and fastened down with mother-of-pearl buttons: plain sleeves, moderately full, and a pelerine cape, with an antique collar of pointed lace falling over.
&
(d)
&
Real, but roughly 1825--1835, and British. Four independent markers, all pointing the same way: pelisse as a current outer-garment name, gros de Naples as a current silk, vandykes as a trimming, pelerine with an antique collar. By the 1870s the pelisse is a costume-history word, gros de Naples has been displaced by gros grain and faille, and the vandyke/pelerine combination is grandmother's wardrobe. ``Cotton-blossom-yellow'' belongs to the \emph{Ackermann's}/\emph{La Belle Assemblée} register of invented poetic colorways. The syntax is right---the verbless nominal opening, the colon splicing the second half---which is exactly why it's a good distractor: it fails on lexical dating, not on style.
\\
\bottomrule
\end{tabular}
\caption{Responses to the assignment: ``Two authentic sentences referencing `sleeves' from a ladies' home magazine published in the United States in the 1870s, and two impostors, either generated by language models or taken from books in other periods. Identify the impostors, and explain your rationale.'' The model's diagnoses are all correct.}
\label{tab:ladies-home-magazine-rationales}
\end{table*}

\section{Fine tuning of candidate models}
\label{sec:appendix-tuning}
To assess whether tuning on period prose improves style judgment and generation, we fine-tuned two models.

\textbf{Qwen 2.5 7B ft.} This was a continued-pretraining run with a small instruction component, not an instruction-tuning run. It drew on 119 volumes drawn from the IDI corpus (1875-1924), sentence-split with NLTK, and packed into ~2,000-token chunks at sentence boundaries. The result was 14.6M training tokens in 18,956 examples. They don't overlap with benchmark books. Note that the date range for this model is not coextensive with the whole benchmark span.

\textbf{GPT-4.1 finetune.} This was pure instruction tuning, not raw-text continuation. 340 training and 25 validation examples, every one a system / user / assistant triple. The corpus is 160 IDI volumes, 16 per decade across 1831–1930, non-overlapping with the benchmark and selected specifically so that no style-judge instrument has ever seen them, with two questions drawn per volume. Question types include cloze, inference, and knowledge/abstention.

\end{document}